\documentclass[10pt,twocolumn,letterpaper]{article}

\usepackage{cvpr}
\usepackage{times}
\usepackage{epsfig}
\usepackage{xspace}
\usepackage{color}
\usepackage{multirow}
\usepackage{graphics}
\usepackage{adjustbox}
\usepackage{subfigure}
\usepackage{amsmath}
\usepackage[noend]{algpseudocode}
\usepackage{algorithmicx,algorithm}
\usepackage{paralist} 
\usepackage{enumitem}
\usepackage{graphicx} 
\usepackage{epstopdf}
\usepackage{booktabs} 

\usepackage[pagebackref=true,breaklinks=true,letterpaper=true,colorlinks,citecolor=blue,linkcolor=blue,bookmarks=false]{hyperref}

\newcommand{\Paragraph}[1]{\noindent\paragraph{#1}} 

\long\def\ignorethis#1{}

\definecolor{Gray}{rgb}{0.35,0.35,0.35}
\definecolor{Blue}{rgb}{0,0.2,0.8}
\definecolor{Red}{rgb}{0.8,0.2,0}
\definecolor{Green}{rgb}{0.0,0.5,0.1}
\definecolor{Gray}{rgb}{0.4,0.4,0.4}

\newlength\paramargin
\newlength\figmargin
\newlength\secmargin

\usepackage{array}
\newcolumntype{L}[1]{>{\raggedright\let\newline\\\arraybackslash\hspace{0pt}}m{#1}}
\newcolumntype{C}[1]{>{\centering\let\newline\\\arraybackslash\hspace{0pt}}m{#1}}
\newcolumntype{R}[1]{>{\raggedleft\let\newline\\\arraybackslash\hspace{0pt}}m{#1}}

\def\ie{i.e.,~}

\def\etal{et~al.\xspace}

\graphicspath{{figure/eps/}}

\cvprfinalcopy 

\def\cvprPaperID{97} 

\ifcvprfinal\fi

\begin{document}
\title{Learning a Discriminative Prior for Blind Image Deblurring}
\author{
    Lerenhan Li$^{1,2}$
    \hspace{25pt}
    Jinshan Pan$^{3}$
    \hspace{25pt}
    Wei-Sheng Lai$^{2}$
    \hspace{25pt}
    Changxin Gao$^{1}$\\
    Nong Sang$^{1}$\thanks{Corresponding author.}
    \hspace{55pt}
    Ming-Hsuan Yang$^{2}$
    \\
    $^1$National Key Laboratory of Science and Technology on Multispectral Information Processing,\\
        School of Automation, Huazhong University of Science and Technology\\
    $^2$Electrical Engineering and Computer Science, University of California, Merced\\
    $^3$School of Computer Science and Engineering, Nanjing University of Science and Technology\\
}
\maketitle

\pagestyle{empty}  
\thispagestyle{empty} 


\begin{abstract}
We present an effective blind image deblurring method based on a data-driven discriminative prior.
Our work is motivated by the fact that a good image prior should favor clear images over blurred ones.
In this work, we formulate the image prior as a binary classifier which can be achieved by a deep convolutional neural network (CNN).
The learned prior is able to distinguish whether an input image is clear or not.
Embedded into the maximum a posterior (MAP) framework, it helps blind deblurring in various scenarios, including natural, face, text, and low-illumination images.
%
However, it is difficult to optimize the deblurring method with the learned image prior as it involves a non-linear CNN.
Therefore, we develop an efficient numerical approach based on the half-quadratic splitting method and gradient decent algorithm to solve the proposed model.
Furthermore, the proposed model can be easily extended to non-uniform deblurring.
Both qualitative and quantitative experimental results show that our method performs favorably against state-of-the-art algorithms as well as domain-specific image deblurring approaches.

\end{abstract}

\section{Introduction}
\label{sec:intro}
Blind image deblurring is a classical problem in image processing and computer vision, which aims to recover a latent image from a blurred input.
When the blur is spatially invariant, the blur process is usually modeled by
\begin{equation}\label{1.1}
    B = I \otimes k + n,
\end{equation}
where $\otimes$ denotes convolution operator, $B, I, k$ and $n$ denote the blurred image, latent sharp image, blur kernel, and noise, respectively.
%
The problem \eqref{1.1} is ill-posed as both $I$ and $k$ are unknown, and there exist infinite solutions.
To tackle this problem, additional constraints and prior knowledge on both blur kernels and images are required.

\begin{figure}[t]
\footnotesize
\centering
\renewcommand{\tabcolsep}{1pt} 
\renewcommand{\arraystretch}{1} 
\begin{center}
\begin{tabular}{cc}
  \includegraphics[width=0.49\linewidth]{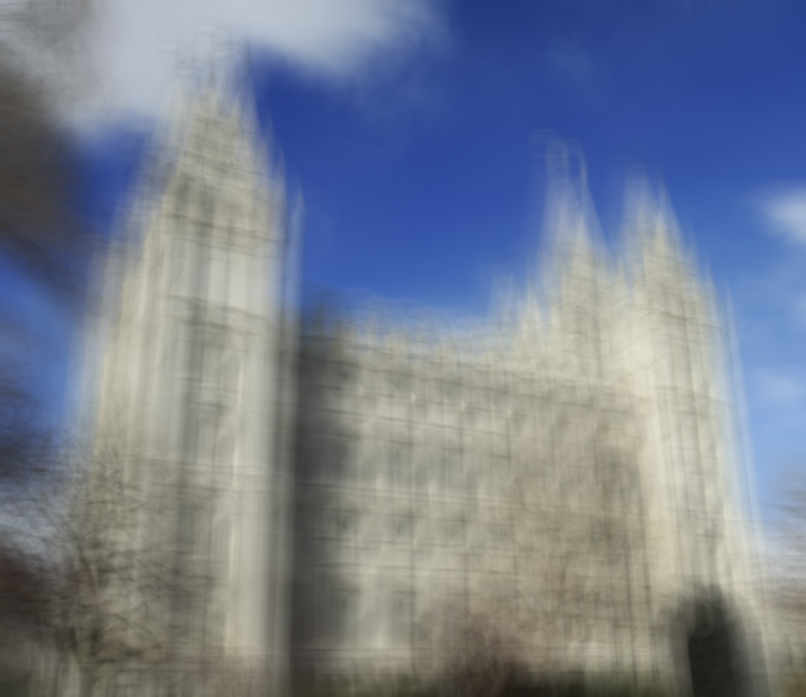} &
  \includegraphics[width=0.49\linewidth]{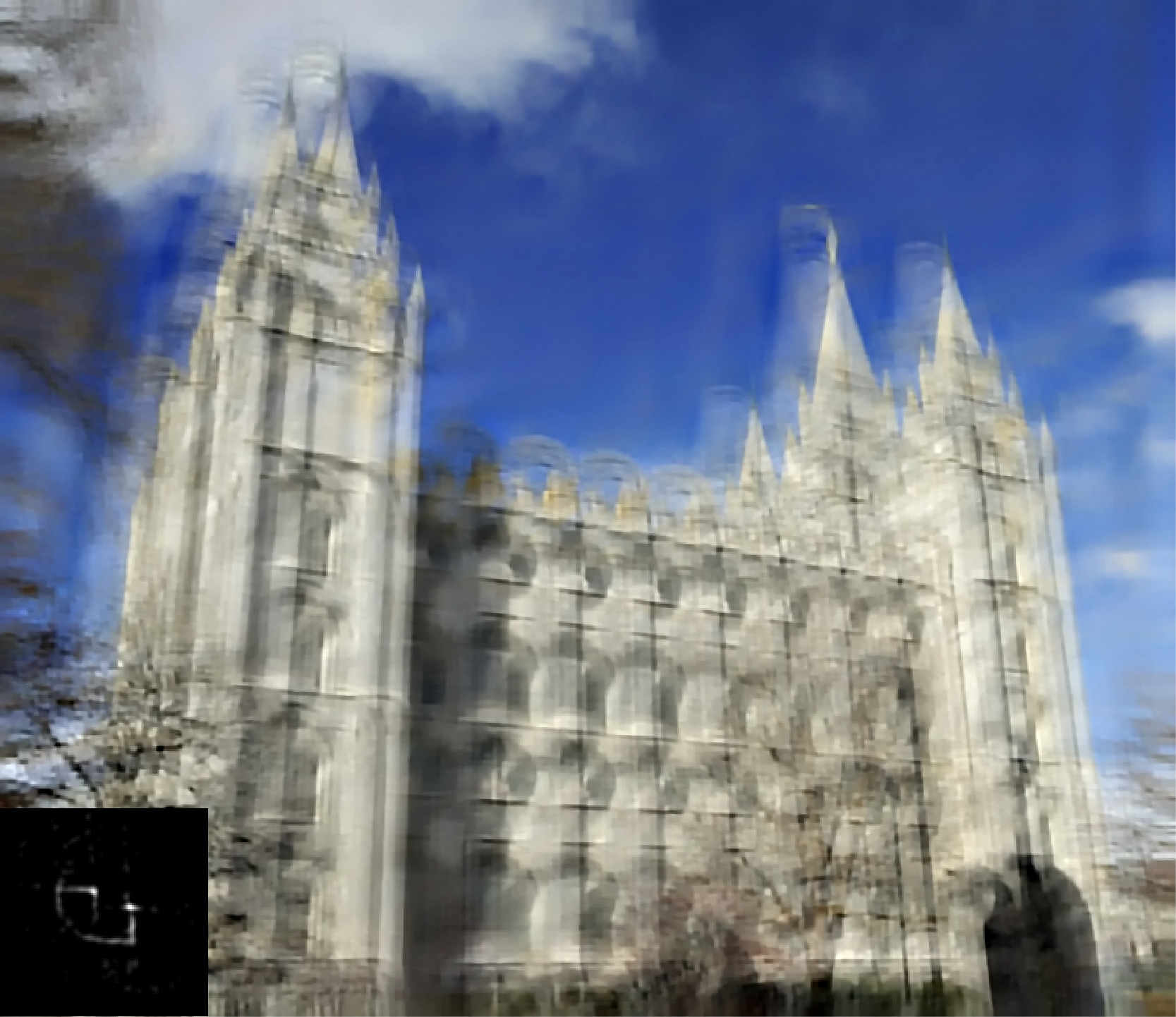} \\
  (a) Blurred image &
  (b) Xu~\etal~\cite{xu2013unnatural} \\
  \includegraphics[width=0.49\linewidth]{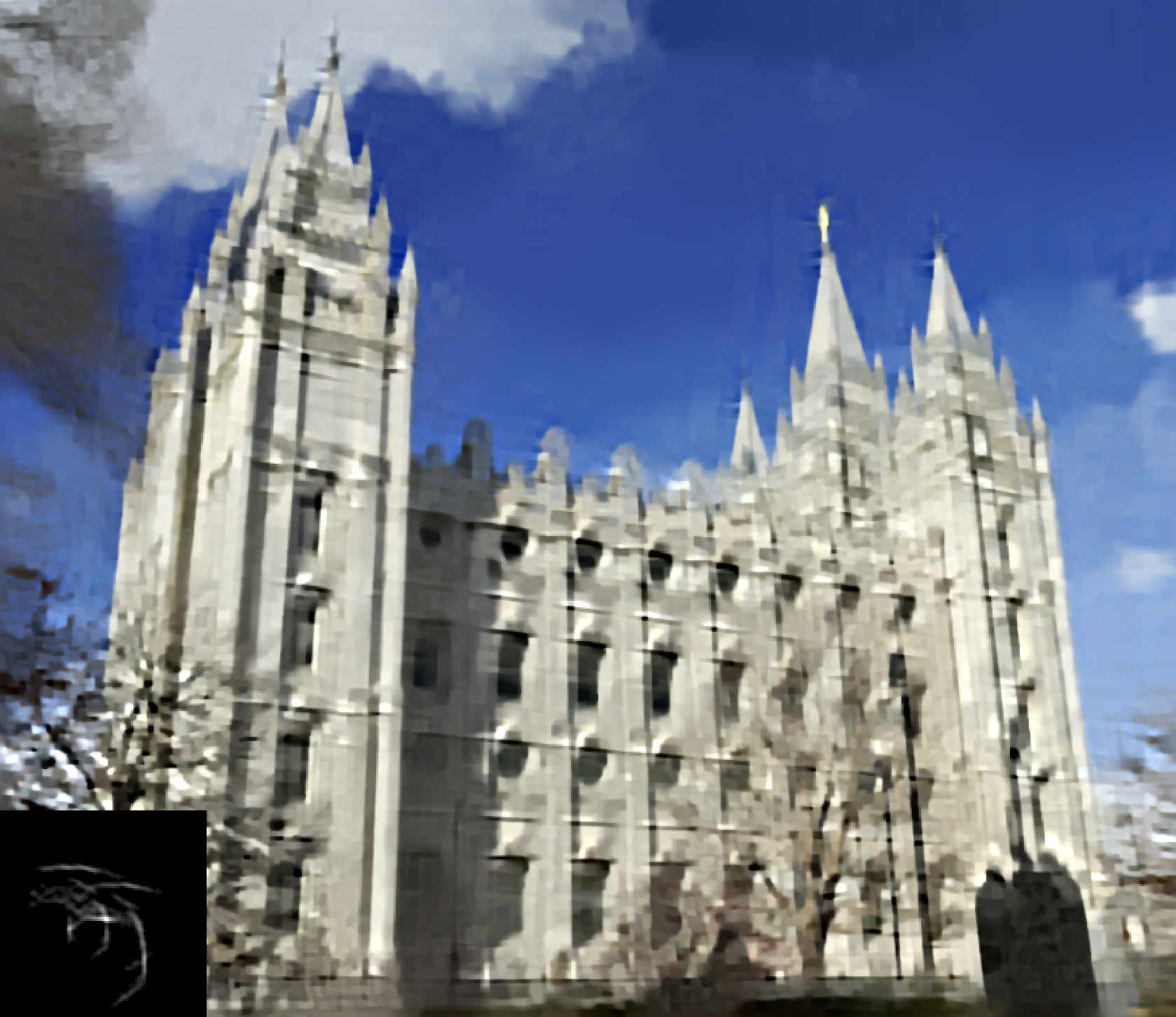} &
  \includegraphics[width=0.49\linewidth]{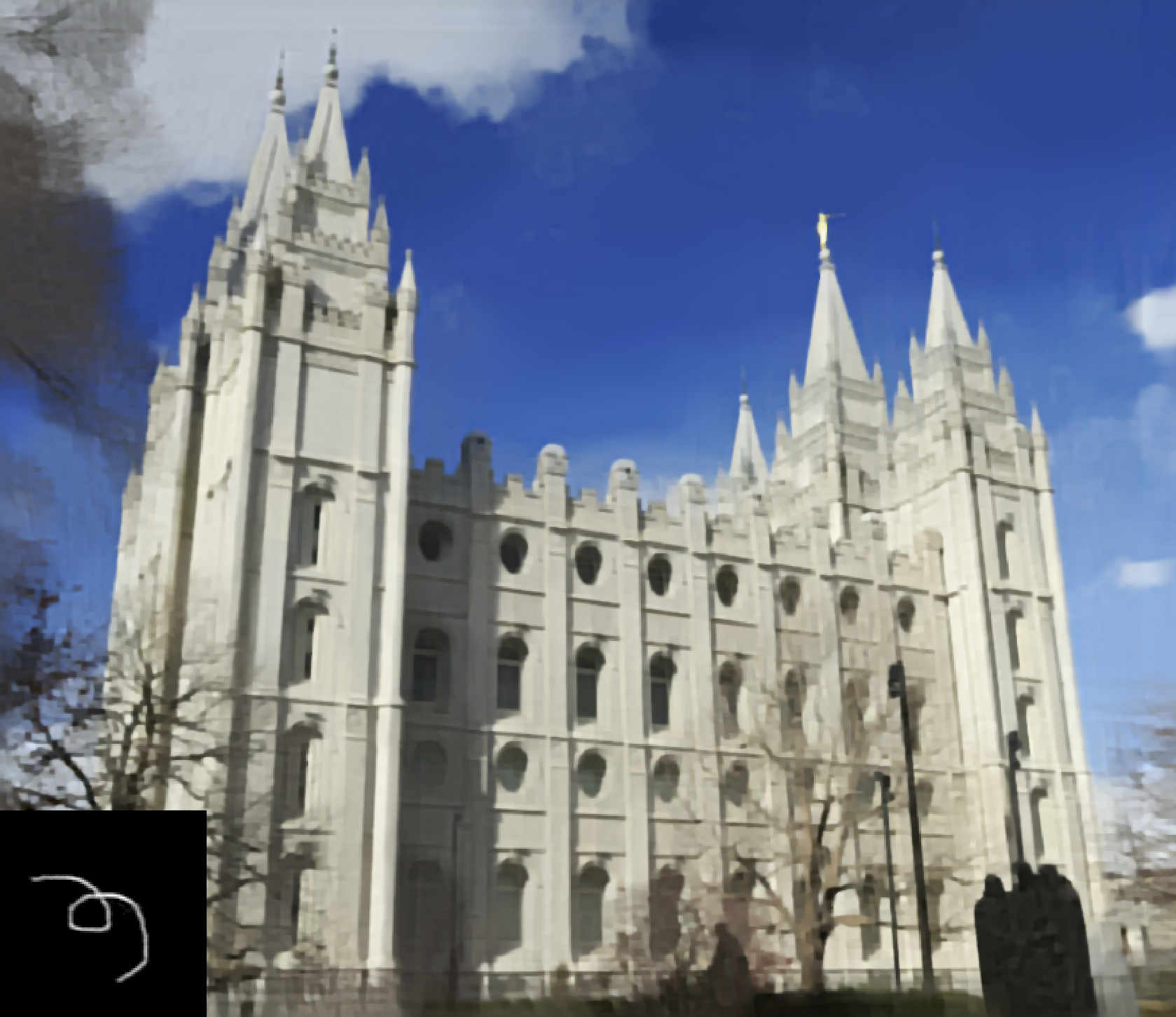} \\
  (c) Pan~\etal~\cite{Pan_2016_CVPR} &
  (d) Ours \\
\end{tabular}
\end{center}
\vspace{-2mm}
\caption{A deblurred example. We propose a discriminative image prior which is learned from a deep binary classification network for image deblurring.
For the blurred image $B$ in (a) and its corresponding clear image $I$, we can get $\frac{{{{\left\| {\nabla I} \right\|}_0}}}{{{{\left\| {\nabla B} \right\|}_0}}} = 0.85$, $\frac{{{{\left\| {D(I)} \right\|}_0}}}{{{{\left\| {D(B)} \right\|}_0}}} = 0.82$ and $\frac{{f(I)}}{{f(B)}} = 0.03$, where $\nabla$, $D( \cdot )$, $\|\cdot\|_0$ and $f( \cdot )$ denote the gradient operator~\cite{xu2013unnatural}, the dark channel~\cite{Pan_2016_CVPR}, $L_0$ norm~\cite{Pan_2016_CVPR,xu2013unnatural} and our proposed classifier, respectively. The prior is more discriminative than the hand-crafted priors, thus leading to better deblurred results. (A larger ratio indicates that the prior responses are closer and cannot be well separated.)}
\vspace{-6mm}
\label{fig:itro}
\end{figure}

The main success of the recent deblurring methods mainly comes from the development of effective image priors and edge-prediction strategies.
However, the edge-prediction based methods usually involve a heuristic edge selection step, which do not perform well when strong edges are not available.
To avoid the heuristic edge selection step, numerous algorithms based on natural image priors have been proposed, including normalized sparsity~\cite{krishnan2011blind}, $L_0$ gradients~\cite{xu2013unnatural} and dark channel prior~\cite{Pan_2016_CVPR}.
%
These algorithms perform well on generic natural images but do not generalize well to specific scenarios, such as text~\cite{Pan2014Deblurring}, face~\cite{panface} and low-illumination images~\cite{hu2014deblurring}.
%
%
Most of the aforementioned image priors have a similar effect that they favor clear images over blurred images, and
this property contributes to the success of the MAP-based methods for blind image deblurring.
However, most priors are hand-crafted and mainly based on limited observations of specific image statistics.
These algorithms cannot be generalized well to handle various scenarios in the wild.
Thus, it is of great interest to develop a general image prior which is able to deal with different scenarios with the MAP framework.

To this end, we formulate the image prior as a binary classifier which is able to distinguish clear images from blurred ones.
Specifically, we first train a deep CNN to classify blurred (labeled as 1) and clear (labeled as 0) images.
%
%
%
To handle arbitrary image sizes in the coarse-to-fine MAP framework, we adopt a global average pooling layer~\cite{lin2013network} in the CNN.
In addition, we use a multi-scale training strategy to make the classifier more robust to different input image sizes.
We then take the learned CNN classifier as a regularization term w.r.t. latent images in the MAP framework.
Figure~\ref{fig:itro} shows an example that the proposed image prior is more discriminative (i.e., has a lower ratio between the response of blurred and clear images) than the state-of-the-art hand-crafted prior~\cite{Pan_2016_CVPR}.

While the intuition behind the proposed method is straightforward, in practice it is difficult to optimize the deblurring method with the learned image prior as a non-linear CNN is involved.
Therefore, we develop an efficient numerical algorithm based on the half-quadratic splitting method and gradient decent approach.
The proposed algorithm converges quickly in practice and can be applied to different scenarios as well as non-uniform deblurring.
%

The contributions of this work are as follows:
\begin{compactitem}
\item We propose an effective discriminative image prior which can be learned by a deep CNN classifier for blind image deblurring.
To ensure that the proposed prior (\ie classifier) can handle the image of different sizes, we use the global average pooling and multi-scale training strategy to train the proposed CNN.
\item We use the learned classifier as a regularization term of the latent image in the MAP framework and develop an efficient optimization algorithm to solve the deblurring model.
\item We demonstrate that the proposed algorithm performs favorably against the state-of-the-art methods on both the widely-used natural image deblurring benchmarks and domain-specific deblurring tasks.
\item We show the proposed method can be directly generalized to the non-uniform deblurring.
\end{compactitem}

\section{Related Work}
\label{sec:related}

Recent years have witnessed significant advances in single image deblurring.
We focus our discussion on recent optimization-based and learning-based methods.

\vspace{-2mm}
\Paragraph{Optimization-based methods.}
State-of-the-art optimization based approaches can be categorized to implicit and explicit edge enhancement methods.
The implicit edge enhancement approaches focus on developing effective image priors to favor clear images over blurred ones.
Representative image priors include sparse gradients~\cite{fergus2006removing,levin2009understanding,xu2010two}, normalized sparsity~\cite{krishnan2011blind}, color-line~\cite{lai2015blur}, $L_0$ gradients~\cite{xu2013unnatural}, patch priors~\cite{sun},  and self-similarity~\cite{michaeli2014blind}.

Although these image priors are effective for deblurring natural images, they are not able to handle specific types of input such as text, face and low-illumination images.
The statistics of these domain-specific images are quite different from natural images.
Thus, Pan et al.~\cite{Pan2014Deblurring} propose the $L_0$-regularized prior on both image intensity and gradients for deblurring text images.
Hu et al.~\cite{hu2014deblurring} detect the light streaks in extremely low-light images for estimating blur kernels.
Recently, Pan~\etal~\cite{Pan_2016_CVPR} propose a dark channel prior for deblurring natural images, which can be applied to face, text and low-illumination images as well.
However, the dark channel prior is less effective when there is no dark pixel in the image.
Yan~\etal~\cite{yan2017image} further propose to incorporate a bright channel prior with the dark channel prior to improve the robustness of the deblurring algorithm.

While those algorithms demonstrate state-of-the-art performance, most priors are hand-crafted and designed under limited observation.
In this work, we propose to learn a data-driven discriminative prior using a deep CNN.
Our prior is designed from a simple criterion without any specific assumption: the prior should favor clear images over blurred images under various of scenarios.

\begin{figure*}[t]
	\footnotesize
	\centering
	\renewcommand{\arraystretch}{1.2} 
	\begin{tabular}{cc}
		\begin{minipage}{0.65\linewidth}
			\includegraphics[width=\linewidth]{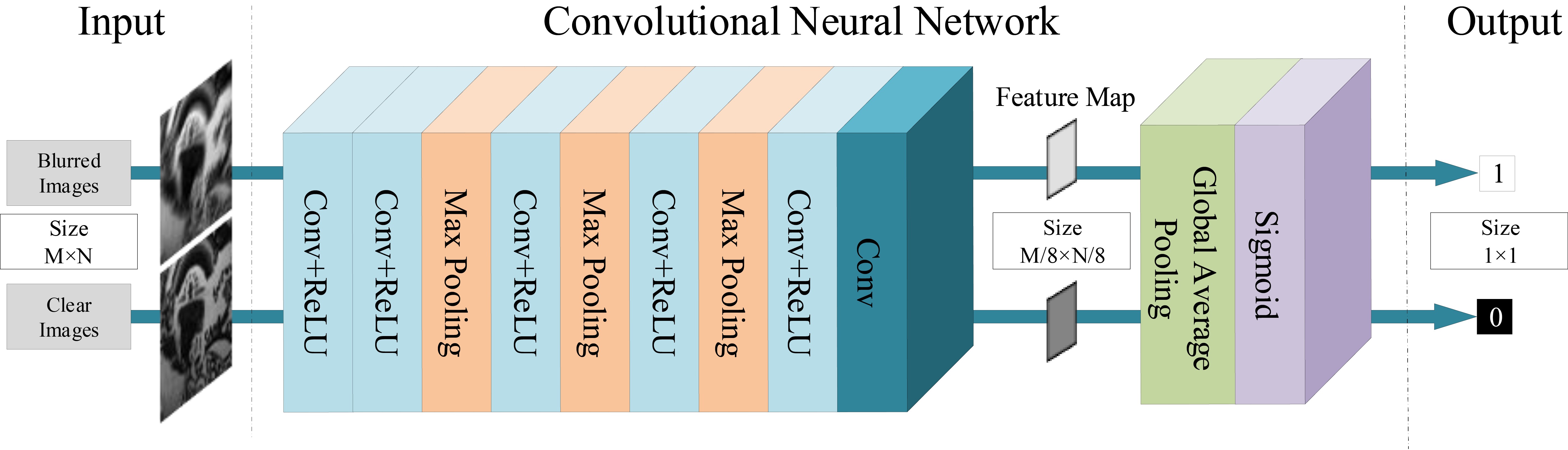}
		\end{minipage}
		&
		{
		\scriptsize
		\begin{tabular}{c|c|c|c}
			\toprule
			Layers & Filter size & Stride & Padding \\
			\midrule
			CR1 & 3$\times$3$\times$1$\times$64  & 1 & 1 \\
			CR2 & 3$\times$3$\times$64$\times$64 & 1 & 1 \\
			M3 & 2$\times$2                    & 2 & 0 \\
			CR4 & 3$\times$3$\times$64$\times$64 & 1 & 1 \\
			M5 & 2$\times$2                    & 2 & 0 \\
			CR6 & 3$\times$3$\times$64$\times$64 & 1 & 1 \\
			M7 & 2$\times$2                    & 2 & 0 \\
			CR8 & 3$\times$3$\times$64$\times$64 & 1 & 1 \\
			C9 & 3$\times$3$\times$64$\times$1  & 1 & 1 \\
			G10 & (M/8)$\times$(N/8)            & 1 & 0 \\
			S11 & - & - & -\\
			\bottomrule
		\end{tabular}
		}
		\\
		(a) Network architecture
		&
		(b) Network parameters
	\end{tabular}	
	\caption{
		Architecture and parameters of the proposed binary classification network.
		We adopt a global average pooling layer instead of a fully-connected layer to handle different sizes of input images.
        CR denotes the convolutional layer followed by a ReLU non-linear function, M denotes the max-pooling layer, C denotes the convolutional layer, G denotes the global average pooling layer and S denotes the sigmoid non-linear function.
	}
	\label{fig:network} 
	\vspace{-4mm}
\end{figure*}

\vspace{-2mm}
\Paragraph{Learning-based methods.}
With the success of deep CNNs on high-level vision problems~\cite{he2016deep,long2015fully}, several approaches have adopted deep CNNs in image restoration problems, including super-resolution~\cite{dong2014learning,VDSR,LapSRN}, denoising~\cite{Mao-NIPS-2016} and JPEG deblocking~\cite{Dong-ICCV-2015}.
Hradi{\v{s}}~\etal~\cite{hradivs2015convolutional} propose an end-to-end CNN to deblur text images.
Following the MAP-based deblurring methods, Schuler~\etal~\cite{schuler2016learning} train a deep network to estimate the blur kernel and then adopt a conventional non-blind deconvolution approach to recover the latent sharp image.
Sun~\etal~\cite{sun2015learning} and Yan and Shao~\cite{yan2016blind} parameterize the blur kernels and learn to estimate them via classification and regression, respectively.
Several approaches train deep CNNs as an image prior or denoiser for non-blind deconvolution~\cite{sreehari2016plug,zhangk2017learning,zhangjw2017learning}, which cannot be directly applied in blind deconvolution.
Recently, Chakrabarti~\cite{chakrabarti2016neural} trains a deep network to predict the Fourier coefficients of a deconvolution filter.
Nevertheless, the performance of deep CNNs on blind image deblurring~\cite{bigdeli2017deep,schelten2015interleaved} still falls behind conventional optimization-based approaches on handling large blur kernels.
In our work, we take advantage of both conventional MAP-based framework and the discriminative ability of deep CNNs.
We embed the learned CNN prior into the coarse-to-fine MAP framework for solving the blind image deblurring problem.

\section{Learning a Data-Driven Image Prior}
\label{sec:learning}
In this section, we describe the motivation of developing the proposed image prior, network design, loss function, and implementation details of our binary classifier.

\subsection{Motivation}

The MAP-based blind image deblurring methods typically solve the following problem:
\begin{equation}\label{deblur_energy_function}
    \mathop {\min }\limits_{I,k} \left\| {I \otimes k - B} \right\|_2^2 + \gamma \left\| k \right\|_2^2 + p(I).
\end{equation}
The key to the success of this framework lies on the latent image prior $p(I)$, which favors clear images over blurred images when minimizing~\eqref{deblur_energy_function}.
Therefore, the image prior $p(I)$ should have lower responses for clear images and higher responses for blurred images.
This observation motivates us to learn a data-driven discriminative prior via binary classification.
We train a deep CNN by predicting blurred images as positive (labeled as 1) and clear images as negative (labeled as 0) samples.
Compared with state-of-the-art latent image priors~\cite{xu2013unnatural,Pan_2016_CVPR}, the assumption of our prior is simple and straightforward without using any hand-crafted functions or assumptions.

\subsection{Binary classification network}
%
Our goal is to train a binary classifier via a deep CNN.
The network takes an image as the input and outputs a single scalar, which represents the probability of the input image to be blurred.
As we aim to embed the network as a prior into the coarse-to-fine MAP framework, the network should be able to handle different sizes of input images.
Therefore, we replace the commonly used fully-connected layers in classifiers with the global average pooling layer~\cite{lin2013network}.
The global average pooling layer converts various sizes of feature maps into a single scalar before the sigmoid layer.
In addition, there is no additional parameter in the global average pooling layer, which alleviates the overfitting problem.
Figure~\ref{fig:network} shows the architecture and detail parameters of our binary classification network.

\subsection{Loss function}
We denote the input image by $x$ and the network parameters to be optimized by $\theta$.
The deep network learns a mapping function $f(x; \theta) = P(x \in \text{Blurred} | x)$ that predicts the probability of the input image to be blurred.
We optimize the network via the binary cross entropy loss function:
\begin{equation}\label{loss}
   L(\theta ) =  - \frac{1}{N}\sum\limits_{i = 1}^N {{{\hat y}_i}\log ({y_i}) + (1 - {{\hat y}_i})\log (1 - {y_i})},
\end{equation}
where $N$ is the number of training samples in a batch, ${y_i} = f({x_i};\theta )$ is the output of the classifier and ${{{\hat y}_i}}$ is the label of the input image.
We assign $\hat{y} = 1$ for blurred images and $\hat{y} = 0$ for clear images.

\subsection{Training details}
We sample 500 clear images from the dataset of Huiskes and Lew~\cite{huiskes2008mir}, including natural, manmade scene, face, low-illumination and text images.
We use the method of Boracchi and Foi~\cite{boracchi2012modeling} to generate 200 random blur kernels with the size ranging from $7 \times 7$ to $51 \times 51$.
We synthesize blurred images by convolving the clear images with blur kernels and adding a Gaussian noise with $\sigma = 0.01$.
We generate a total of 100,000 blurred images for training.
During training, we randomly crop $200 \times 200$ patches from the training images.
In order to make the classifier more robust to different sizes of images, we adopt a multi-scale training strategy by randomly resizing the input images between $[0.25, 1]$.

We implement the network using the MatConvNet~\cite{vedaldi2015matconvnet} toolbox.
We use the Xavier method to initialize the network parameters and use the Stochastic Gradient Descent (SGD) method for optimizing the network.
We use the batch size of 50, the momentum of 0.9 and the weight decay of ${10^{ - 4}}$.
The learning rate is set to 0.001 and decreased by a factor of 5 for every 50 epochs.

\section{Blind Image Deblurring}
\label{sec:optimization}
After the training process of the proposed network converges, we use the trained model as the latent image prior $p(\cdot)$ in~\eqref{deblur_energy_function}.
In addition, we use the $L_0$ gradient prior~\cite{xu2013unnatural,Pan_2016_CVPR} as a regularization term.
Therefore, we aim to solve the following optimization problem:
\begin{equation}\label{4.1}
    \mathop {\min }\limits_{I,k} \left\| {I \otimes k - B} \right\|_2^2 + \gamma \left\| k \right\|_2^2 + \mu {\left\| {\nabla I} \right\|_0} + \lambda f(I),
\end{equation}
where $\gamma$, $\mu$ and $\lambda$ are the hyper-parameters to balance the weight of each term.

We optimize~\eqref{4.1} by solving the latent image $I$ and the blur kernel $k$ alternatively.
Thus, we divide the problem into $I$ sub-problem:
\begin{equation}\label{4.2}
    \mathop {\min }\limits_I \left\| {I \otimes k - B} \right\|_2^2 + \mu {\left\| {\nabla I} \right\|_0} + \lambda f(I),
\end{equation}
and $k$ sub-problem:
\begin{equation}\label{4.3}
    \mathop {\min }\limits_k \left\| {I \otimes k - B} \right\|_2^2 + \gamma \left\| k \right\|_2^2.
\end{equation}

\subsection{Solving $I$}
\label{sec:I-sub}
In~\eqref{4.2}, both $f( \cdot )$ and ${\left\| {\nabla I} \right\|_0}$ are highly non-convex, which make minimizing~\eqref{4.2} computationally intractable.
To tackle this issue, we adopt the half-quadratic splitting method~\cite{xu2011image} by introducing the auxiliary variables $u$ and $g = ({g_h},{g_v})$ with respect to the image and its gradients in horizonal and vertical directions, respectively.
The energy function~\eqref{4.2} can be rewritten as
\begin{equation}\label{4.4}
    \begin{array}{l}
\mathop {\min }\limits_{I,g,u} \left\| {I \otimes k - B} \right\|_2^2 + \alpha \left\| {\nabla I - g} \right\|_2^2\\
{\kern 1pt} {\kern 1pt} {\kern 1pt} {\kern 1pt} {\kern 1pt} {\kern 1pt} {\kern 1pt} {\kern 1pt} {\kern 1pt} {\kern 1pt} {\kern 1pt} {\kern 1pt} {\kern 1pt} {\kern 1pt} {\kern 1pt} {\kern 1pt} {\kern 1pt} {\kern 1pt} {\kern 1pt} {\kern 1pt} {\kern 1pt} {\kern 1pt}  + \beta \left\| {I - u} \right\|_2^2{\kern 1pt}  + \mu {\left\| g \right\|_0} + \lambda f(u)
\end{array},
\end{equation}
where $\alpha$ and $\beta$ are penalty parameters.
When $\alpha$ and $\beta$ approach infinity, the solution of (\ref{4.4}) is equivalent to that of (\ref{4.2}).
We can solve (\ref{4.4}) by minimizing $I$, $g$ and $u$ alternatively and thus avoid directly minimizing the non-convex functions $f( \cdot )$ and ${\left\| {\nabla I} \right\|_0}$.

We solve the latent image $I$ by fixing $g$ and $u$ and optimizing:
\begin{equation}\label{4.5}
    \mathop {\min }\limits_{I} \left\| {I \otimes k - B} \right\|_2^2 + \alpha \left\| {\nabla I - g} \right\|_2^2 + \beta \left\| {I - u} \right\|_2^2{\kern 1pt},
\end{equation}
which is a least squares optimization and has a closed-form solution:
\begin{equation}\label{4.6}
\scriptsize    I = {F^{ - 1}}\left( {\frac{{\overline {F(k)} F(B) + \beta F(u) + \alpha \left( {\sum\nolimits_{d \in \left\{ {h,v} \right\}} {\overline {F({\nabla _d})} F({g_d})} } \right)}}{{\overline {F(k)} F(k) + \beta  + \alpha \left( {\sum\nolimits_{d \in \left\{ {h,v} \right\}} {\overline {F({\nabla _d})} F({\nabla _d})} } \right)}}} \right),
\end{equation}
where $F( \cdot )$ and ${F^{ - 1}}( \cdot )$ denote the Fourier and inverse Fourier transforms;
$\overline {F( \cdot )}$ is the complex conjugate operator;
${{\nabla _h}}$ and ${{\nabla _v}}$ are the horizontal and vertical differential operators, respectively.

Given the latent image $I$, we solve $g$ and $u$ by:
\begin{align}
    \mathop {\min }\limits_g \alpha \left\| {\nabla I - g} \right\|_2^2 + \mu {\left\| g \right\|_0}, \label{4.7}
    \\
    \mathop {\min }\limits_u \beta \left\| {I - u} \right\|_2^2 + \lambda f(u).\label{4.8}
\end{align}
We solve (\ref{4.7}) following the strategy of Pan et al.~\cite{Pan2014Deblurring} and use the back-propagation approach to compute the derivative of $f( \cdot )$.
We update $u$ using the gradient descent method:
\begin{equation}\label{4.9}
    {{{u}}^{(s + 1)}} = {{{u}}^{(s)}} - \eta \left[ {\beta \left( {{{{u}}^{(s)}} - {{I}}} \right) + \lambda \frac{{df({{{u}}^{(s)}})}}{{d{{{u}}^{(s)}}}}} \right],
\end{equation}
where $\eta$ is the step size.
We summarize the main steps for solving (\ref{4.9}) in Algorithm~\ref{alg:1}.
\begin{algorithm}[!t]
	\footnotesize
\caption{\label{alg:1}Solving (\ref{4.9})}
\hspace*{0.02in} {\bf Input:}
Latent Image $I$\\
\hspace*{0.02in} {\bf Output:}
the solution of $u$.
\begin{algorithmic}[1]
\State initialize ${{u}^{(0)}} \leftarrow {I}$
\While{$s < {s_{\max }}$}
    \State solve for ${{u}^{(s + 1)}}$ by (\ref{4.9}).
    \State $s \leftarrow s + 1$
\EndWhile
\State {\bf{end while}}
\end{algorithmic}
\end{algorithm}


\subsection{Solving $k$}
\label{sec:k-sub}
%
%
In order to obtain more accurate results, we estimate the blur kernel using image gradients~\cite{cho2009fast, Pan2014Deblurring, Pan_2016_CVPR}:
\begin{equation}\label{4.10}
    \mathop {\min }\limits_k \left\| {\nabla I \otimes k - \nabla B} \right\|_2^2 + \gamma \left\| k \right\|_2^2,
\end{equation}
which can also be efficiently solved by the Fast Fourier Transform (FFT).
We then set the negative elements in $k$ to 0 and normalize $k$ so that the sum of all elements is equal to 1.
%
%
We use the coarse-to-fine strategy with an image pyramid~\cite{Pan2014Deblurring, Pan_2016_CVPR} to optimize~\eqref{4.1}.
At each pyramid level, we alternatively solve~\eqref{4.2} and~\eqref{4.10} with $\text{iter}_{\max}$ iterations.
The main steps are summarized in supplemental materials.

\section{Extension to Non-Uniform Deblurring}
\label{sec:nonuniform}
The proposed discriminative image prior can be easily extended for non-uniform motion deblurring.
Based on the geometric model of camera motion~\cite{tai2011richardson, whyte2012non}, we represent the blurred images as the weighted sum of a latent clear image under geometry transformations:
\begin{equation}\label{5.1}
    {\bf{B}} = \sum\limits_t {{k_t}{{\bf{H}}_t}{\bf{I}}}  + {\bf{n}},
\end{equation}
where ${\bf{B}}$, ${\bf{I}}$ and ${\bf{n}}$ are the blurred image, latent image and noise in the vector forms, respectively;
$t$ denotes the index of camera pose samples;
${{k_t}}$ is the weight of the $t$-th camera pose satisfying ${k_t} \ge 0$, $\sum_t {{k_t} = 1}$;
${\bf{H}}_t$ denotes a matrix derived from the homography~\cite{whyte2012non}.
We use the bilinear interpolation when applying ${\bf{H}}_t$ on a latent image ${\bf{I}}$.
Therefore, we simplify (\ref{5.1}) to:
\begin{equation}\label{5.2}
    {\bf{B}} = {\bf{KI}} + {\bf{n}} = {\bf{Ak}} + {\bf{n}},
\end{equation}
where ${\bf{K}} = \sum\nolimits_t {{k_t}{{\bf{H}}_t}}$, $A = [{{\bf{H}}_1}{\bf{I}},{\kern 1pt}  {\kern 1pt} {{\bf{H}}_2}{\bf{I}},{\kern 1pt} {\kern 1pt}  {\kern 1pt}  \ldots ,{\kern 1pt} {\kern 1pt}  {{\bf{H}}_t}{\bf{I}}]$ and ${\bf{k}} = {[{k_1},{\kern 1pt} {\kern 1pt} {k_2},{\kern 1pt} {\kern 1pt} {\kern 1pt}    \ldots ,{\kern 1pt} {\kern 1pt}  {k_t}]^T}$.
We solve the non-uniform deblurring problem by alternatively minimizing:
\begin{equation}\label{5.3}
    \mathop {\min }\limits_{\bf{I}} \left\| {{\bf{KI}} - {\bf{B}}} \right\|_2^2 + \lambda f({\bf{I}}) + \mu {\left\| {\nabla {\bf{I}}} \right\|_0}
\end{equation}
and
\begin{equation}\label{5.4}
    \mathop {\min }\limits_{\bf{k}} \left\| {{\bf{Ak}} - {\bf{B}}} \right\|_2^2 + \gamma \left\| {\bf{k}} \right\|_2^2.
\end{equation}
The optimization methods of (\ref{5.3}) and (\ref{5.4}) are similar to those used for solving \eqref{4.2} and \eqref{4.3}.
The latent image ${\bf{I}}$ and the weight ${\bf{k}}$ are estimated by the fast forward approximation~\cite{hirsch2011fast}.


\begin{figure}[!t]
\footnotesize
\centering
\renewcommand{\tabcolsep}{1pt} 
\renewcommand{\arraystretch}{1} 
\begin{center}
\begin{tabular}{cc}
  \includegraphics[width=0.52\linewidth]{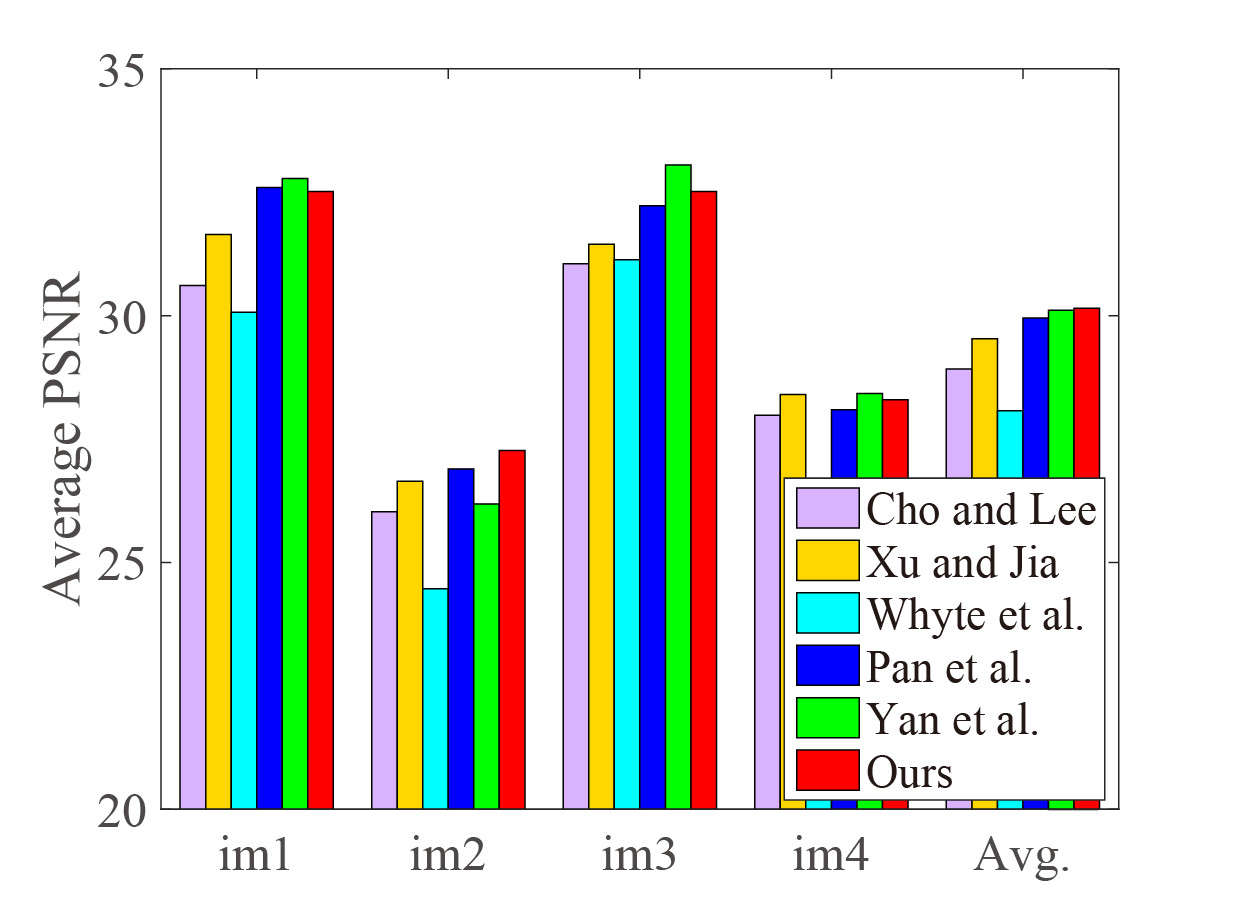} &
  \includegraphics[width=0.48\linewidth]{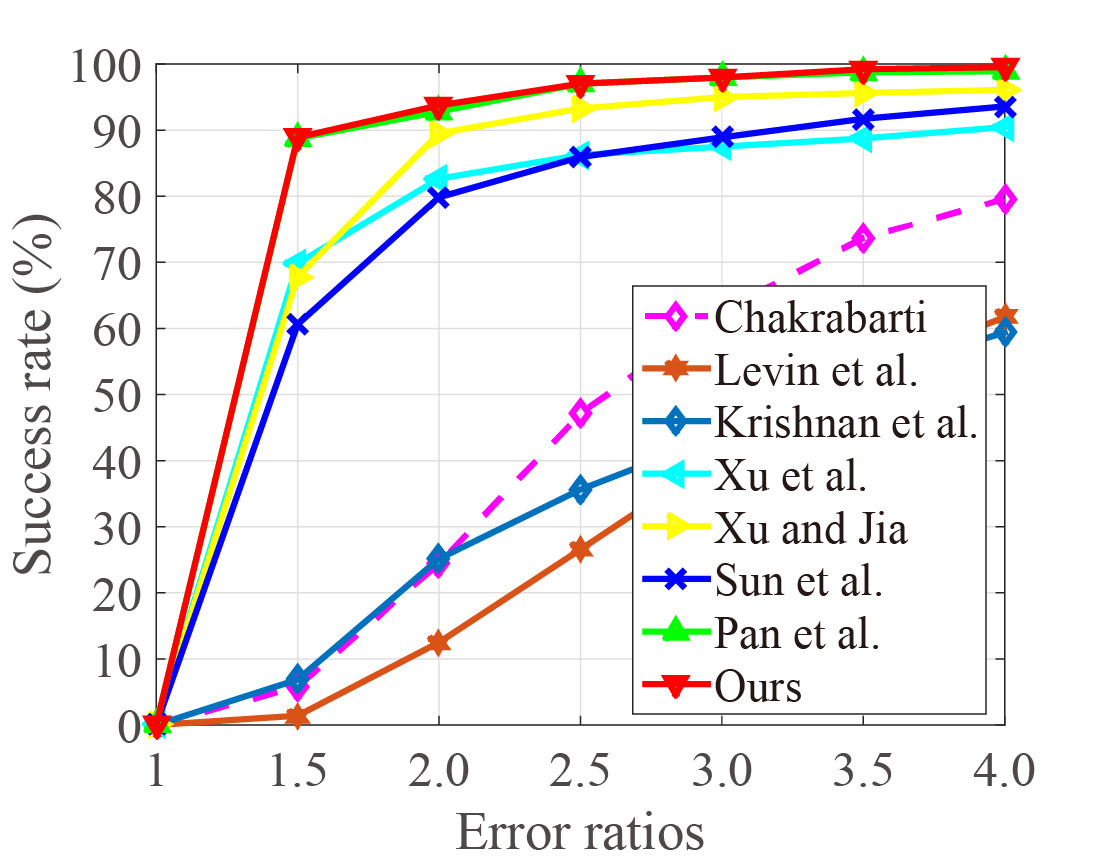} \\
  (a) Results on dataset~\cite{Kohler} & (b) Results on dataset~\cite{sun} \\
\end{tabular}
\end{center}
\vspace{-3mm}
\caption{Quantitative evaluations on benchmark datasets~\cite{Kohler} and~\cite{sun}.}
\label{fig:benchmark} 
\end{figure}

\begin{figure*}[!t]
\footnotesize
\centering
\renewcommand{\tabcolsep}{1pt} 
\renewcommand{\arraystretch}{1} 
\begin{center}
\begin{tabular}{ccccc}
  \includegraphics[width=0.195\linewidth]{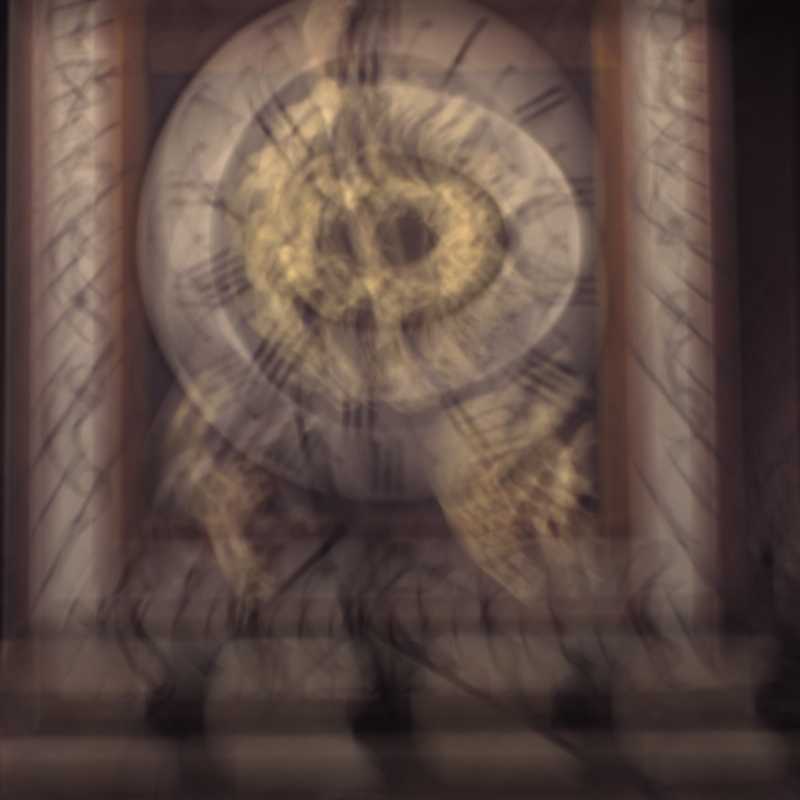} & \includegraphics[width=0.195\linewidth]{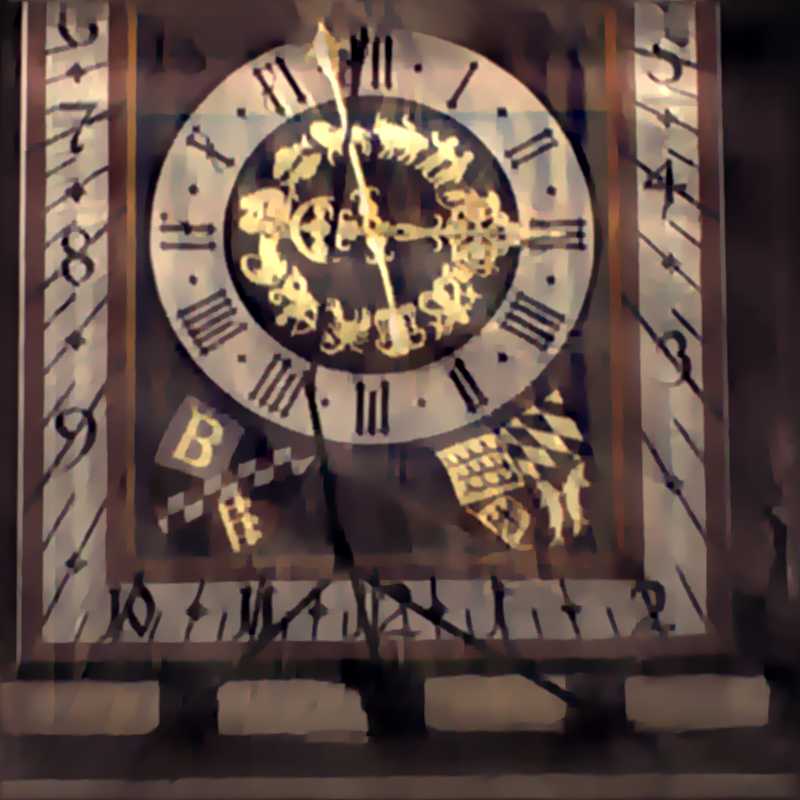}
  & \includegraphics[width=0.195\linewidth]{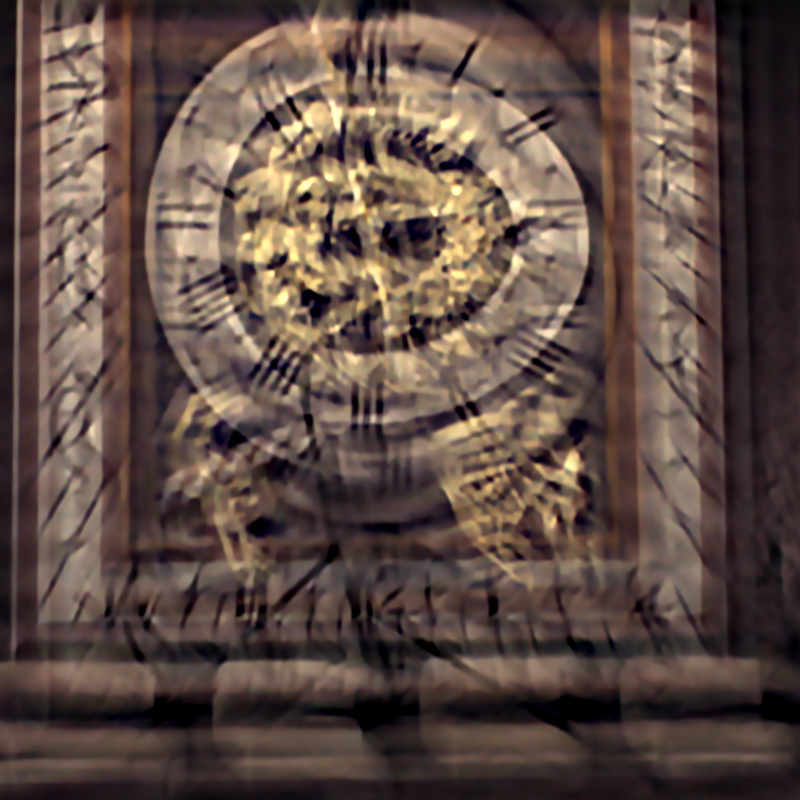}
  & \includegraphics[width=0.195\linewidth]{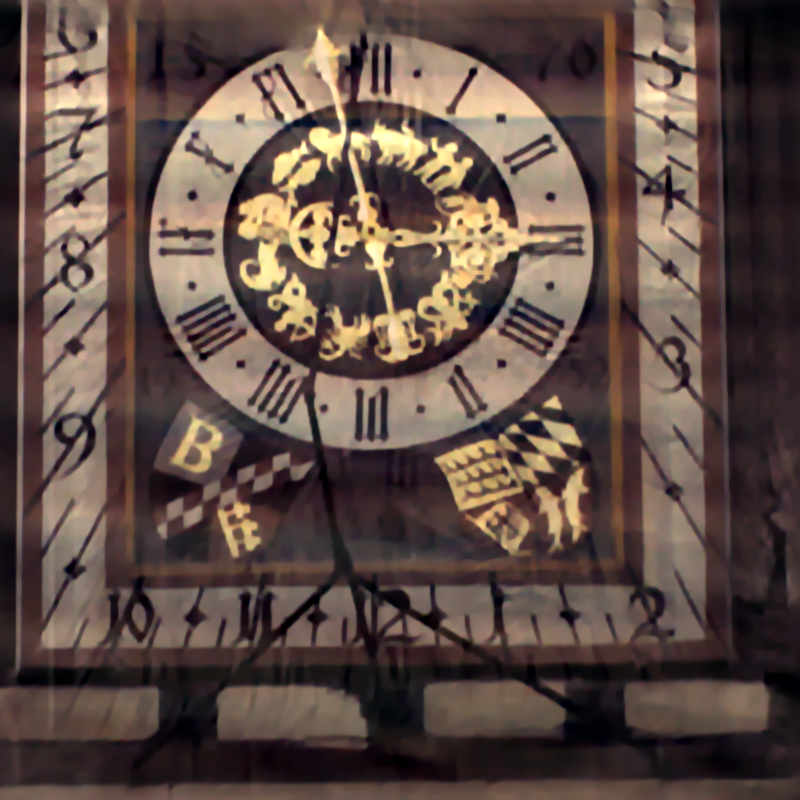}
  & \includegraphics[width=0.195\linewidth]{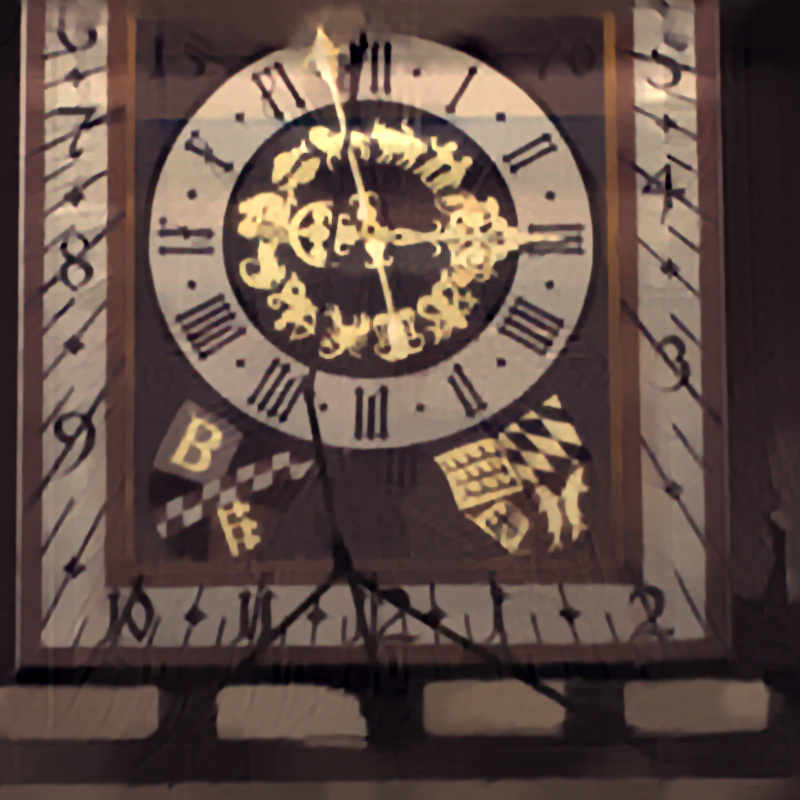}\\
  (a) Blurred image
  & (b) Cho and Lee~\cite{cho2009fast}
  & (c) Yan~\etal~\cite{yan2017image}
  & (d) Pan~\etal~\cite{Pan_2016_CVPR}
  & (e) Ours\\
\end{tabular}
\end{center}
\vspace{-3mm}
\caption{
	A challenging example from dataset~\cite{Kohler}.
	The proposed algorithm restores more visually pleasing results with less ringing artifacts.}
\label{fig:eccv} 
\end{figure*}

\section{Experimental Results}
\label{sec:results}
We evaluate the proposed algorithm on natural image datasets~\cite{Kohler,sun} as well as text~\cite{Pan2014Deblurring}, face~\cite{panface}, and low-illumination~\cite{hu2014deblurring} images.
In all the experiments, we set $\lambda {\rm{ = }}\mu {\rm{ = }}0.004$, $\gamma {\rm{ = }}2$, and $\eta {\rm{ = }}0.1$.
To balance the accuracy and speed, we empirically set $\text{iter}_{\max} = 5$ and ${s_{\max }} = 10$.
Unless specially mentioned, we use the non-blind method in~\cite{Pan2014Deblurring} to recover the final latent images after estimating blur kernels.
All the experiments are carried out on a desktop computer with an Intel Core i7-3770 processor and 32 GB RAM.
The source code and the datasets used in the paper are publicly available on the authors' websites.
More experimental results are included in supplemental material.

\subsection{Natural images}
We first evaluate the proposed algorithm on the natural image dataset of K{\"o}hler \etal~\cite{Kohler}, which contains 4 latent images and 12 blur kernels.
We compare with the 5 generic image deblurring methods~\cite{cho2009fast, xu2010two, whyte2012non, Pan_2016_CVPR, yan2017image}.
We follow the protocol of~\cite{Kohler} to compute the PSNR by comparing each restored image with 199 clear images captured along the same camera motion trajectory.
As shown in Figure~\ref{fig:benchmark} (a), our method achieves the highest PSNR on average.
Figure~\ref{fig:eccv} shows the deblurred results of one example.
Our method generates clearer images with less ringing artifacts.

Next, we evaluate our algorithm on the dataset provided by Sun~\etal~\cite{sun}, which consists of 80 clear images and 8 blur kernels from Levin~\etal~\cite{levin2009understanding}.
We compare with the 6 optimization-based deblurring methods~\cite{levin2011efficient, krishnan2011blind, xu2013unnatural, xu2010two, sun, Pan_2016_CVPR} (solid curves) and one learning-based method~\cite{chakrabarti2016neural} (dotted curve).
For fair comparisons, we apply the same non-blind deconvolution~\cite{zoran2011learning} to restore the latent images.
We measure the error ratio~\cite{levin2009understanding} and plot the results in Figure~\ref{fig:benchmark} (b), which demonstrates that the proposed method performs competitively against the state-of-the-art algorithms.

\begin{figure}
\footnotesize
\centering
\renewcommand{\tabcolsep}{1pt} 
\renewcommand{\arraystretch}{1} 
\begin{center}
\begin{tabular}{ccc}
  \includegraphics[width=0.32\linewidth]{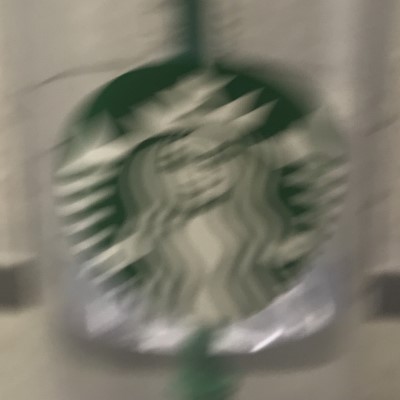}& \includegraphics[width=0.32\linewidth]{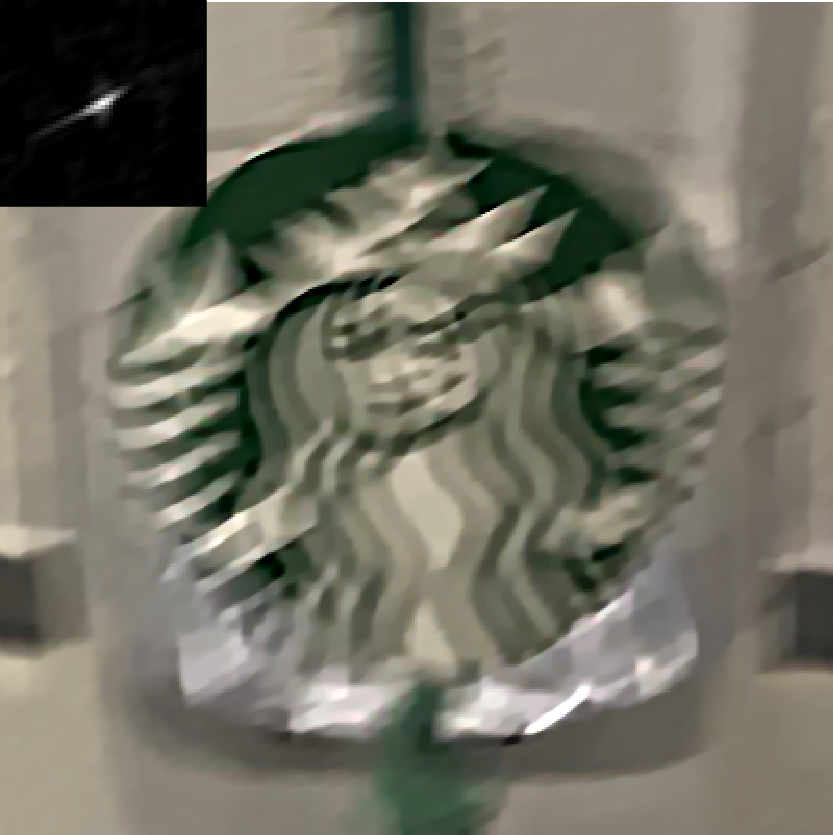}&
  \includegraphics[width=0.32\linewidth]{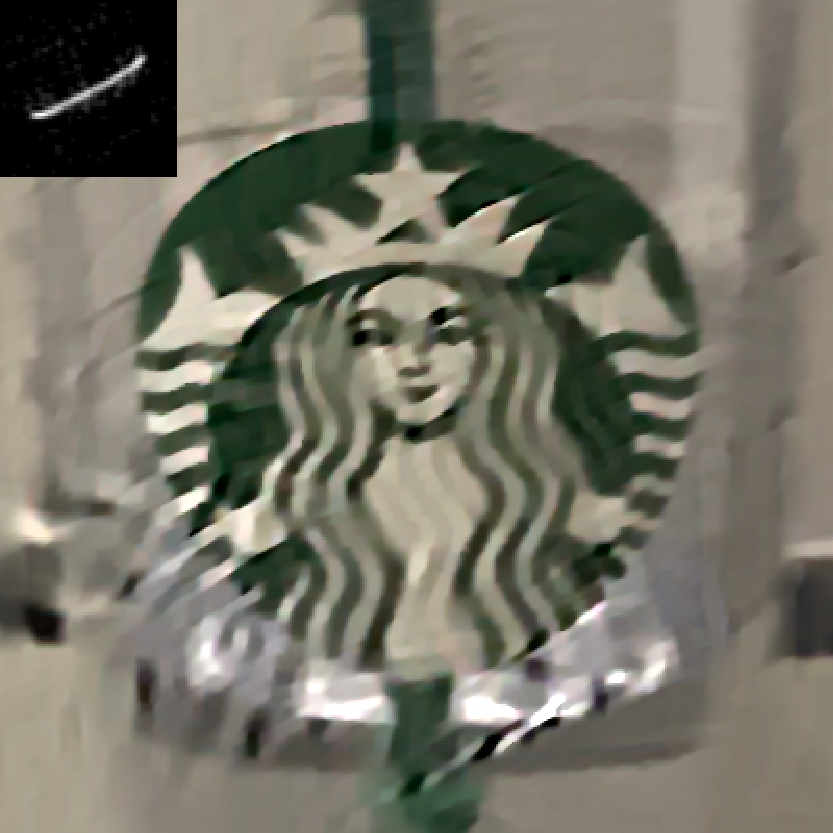}\\
  (a) Blurred image& (b) Krishnan~\etal~\cite{krishnan2011blind} & (c) Xu~\etal~\cite{xu2013unnatural}\\
  \includegraphics[width=0.32\linewidth]{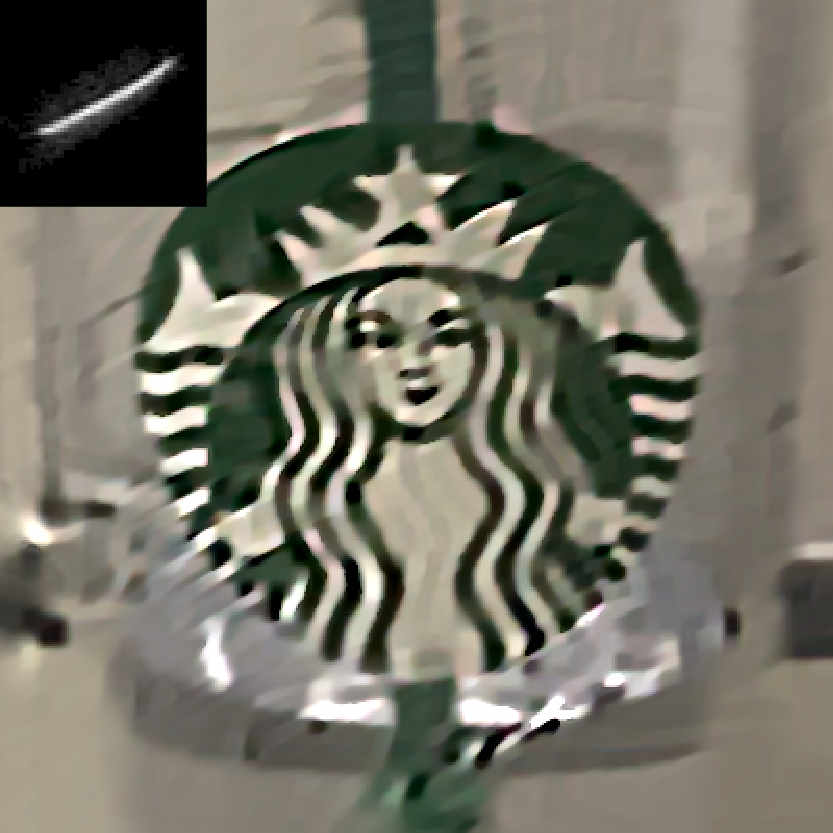}& \includegraphics[width=0.32\linewidth]{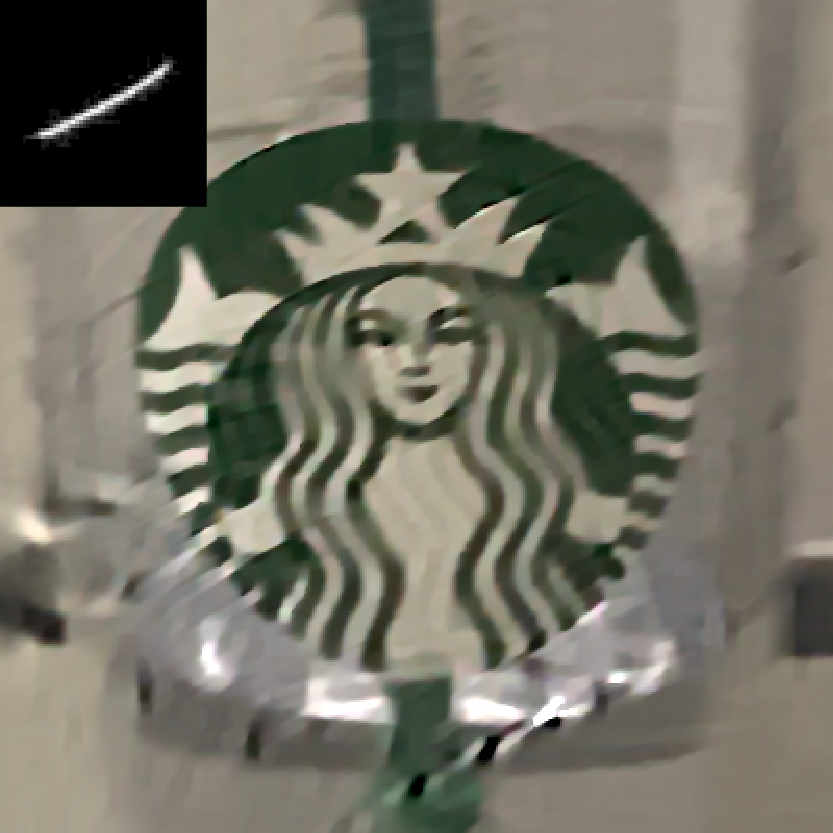}&
  \includegraphics[width=0.32\linewidth]{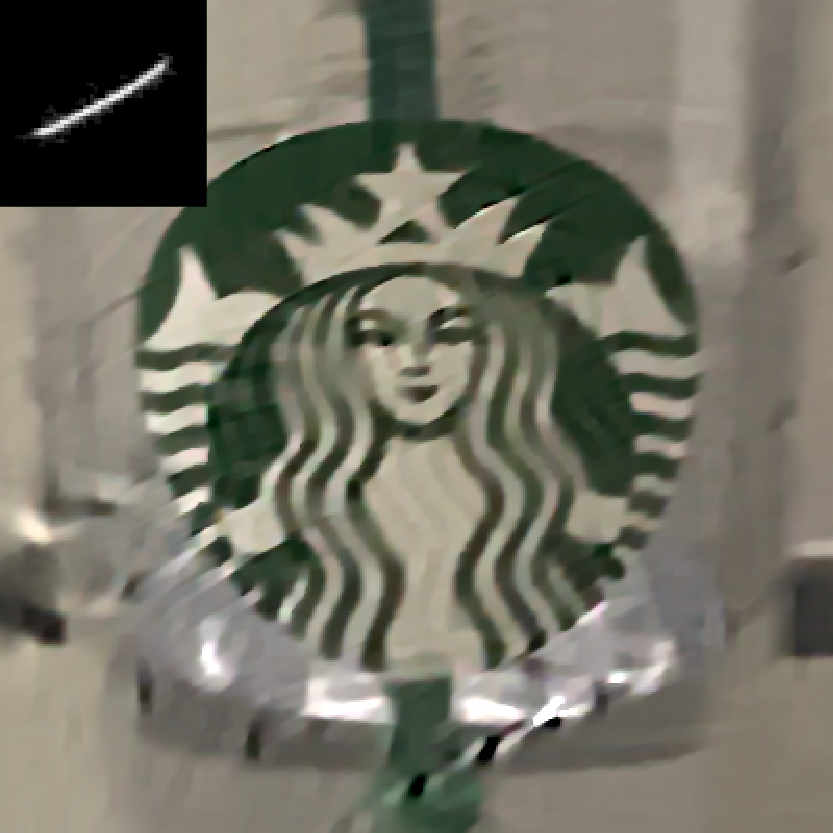}\\
  (d) Pan~\etal~\cite{Pan2014Deblurring}& (d) Pan~\etal~\cite{Pan_2016_CVPR}& (e) Ours\\
\end{tabular}
\end{center}
\caption{Deblurred results on a real blurred image. Our result is sharper with less artifacts.}
\label{fig:real}
\end{figure}

We also test our method on real-world blurred images.
Here we use the same non-blind deconvolution algorithm~\cite{Pan2014Deblurring} for fair comparisons.
As shown in Figure~\ref{fig:real}, our method generates clearer images with fewer artifacts compared with the methods~\cite{krishnan2011blind, xu2013unnatural, Pan2014Deblurring}.
And our result is comparable to the method~\cite{Pan_2016_CVPR}.
%
%
%

\subsection{Domain-specific images}

\begin{figure}
\footnotesize
\centering
\renewcommand{\tabcolsep}{1pt} 
\renewcommand{\arraystretch}{1} 
\begin{center}
\begin{tabular}{cccc}
  \includegraphics[width=0.24\linewidth]{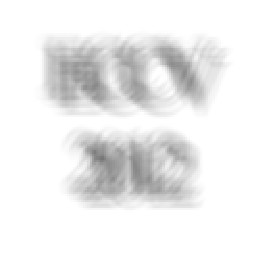} &  \includegraphics[width=0.24\linewidth]{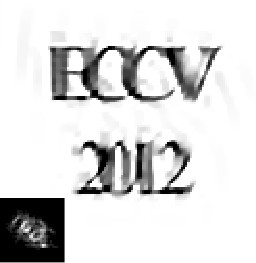} &  \includegraphics[width=0.24\linewidth]{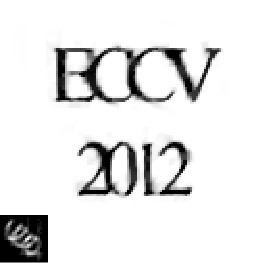} &  \includegraphics[width=0.24\linewidth]{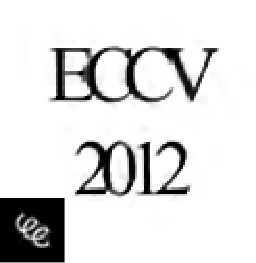} \\
  (a) Blurred image &  (b) Pan~\etal~\cite{Pan_2016_CVPR} &  (c) Pan~\etal~\cite{Pan2014Deblurring} &  (d) Ours\\
\end{tabular}
\end{center}
\vspace{-2mm}
\caption{Deblurred results on a text image. Our method produces sharper deblurred image with more clearer characters than the state-of-the-art text deblurring algorithm~\cite{Pan2014Deblurring}. }
\label{fig:text} 
\end{figure}

We evaluate our algorithm on the text image dataset~\cite{Pan2014Deblurring}, which consists of 15 clear text images and 8 blur kernels from Levin~\etal~\cite{levin2009understanding}.
We show the average PSNR in Table~\ref{tab:1}.
Although the text deblurring approach~\cite{Pan2014Deblurring} has the highest PSNR, the proposed method performs favorably against state-of-the-art generic deblurring algorithms~\cite{cho2009fast, xu2010two, levin2011efficient, xu2013unnatural, Pan_2016_CVPR}.
Figure~\ref{fig:text} shows the deblurred results on a blurred text image.
The proposed method generates much sharper results with clearer characters.

\begin{table}
	\centering
	\caption{
		Quantitative evaluations on text image dataset~\cite{Pan2014Deblurring}.
		Our method performs favorably against generic image deblurring approaches and is comparable to the text deblurring method~\cite{Pan2014Deblurring}.
	}
    \label{tab:1}
	\footnotesize
	\vspace{1mm}
	\begin{tabular}{c|c}
		\toprule
		Methods & Average PSNRs \\
		\midrule
		Cho and Lee~\cite{cho2009fast} & 23.80 \\
		Xu and Jia~\cite{xu2010two} & 26.21 \\
		Levin \etal~\cite{levin2011efficient} & 24.90 \\
		Xu \etal~\cite{xu2013unnatural} & 26.21 \\
		Pan \etal~\cite{Pan_2016_CVPR} (Dark channel) & 27.94 \\
		Pan \etal~\cite{Pan2014Deblurring} (Text deblurring) & 28.80 \\
		Ours & 28.10 \\
		\bottomrule
	\end{tabular}
\end{table}

\begin{figure}
\footnotesize
\centering
\renewcommand{\tabcolsep}{1pt} 
\renewcommand{\arraystretch}{1} 
\begin{center}
\begin{tabular}{cc}
  \includegraphics[width=0.49\linewidth]{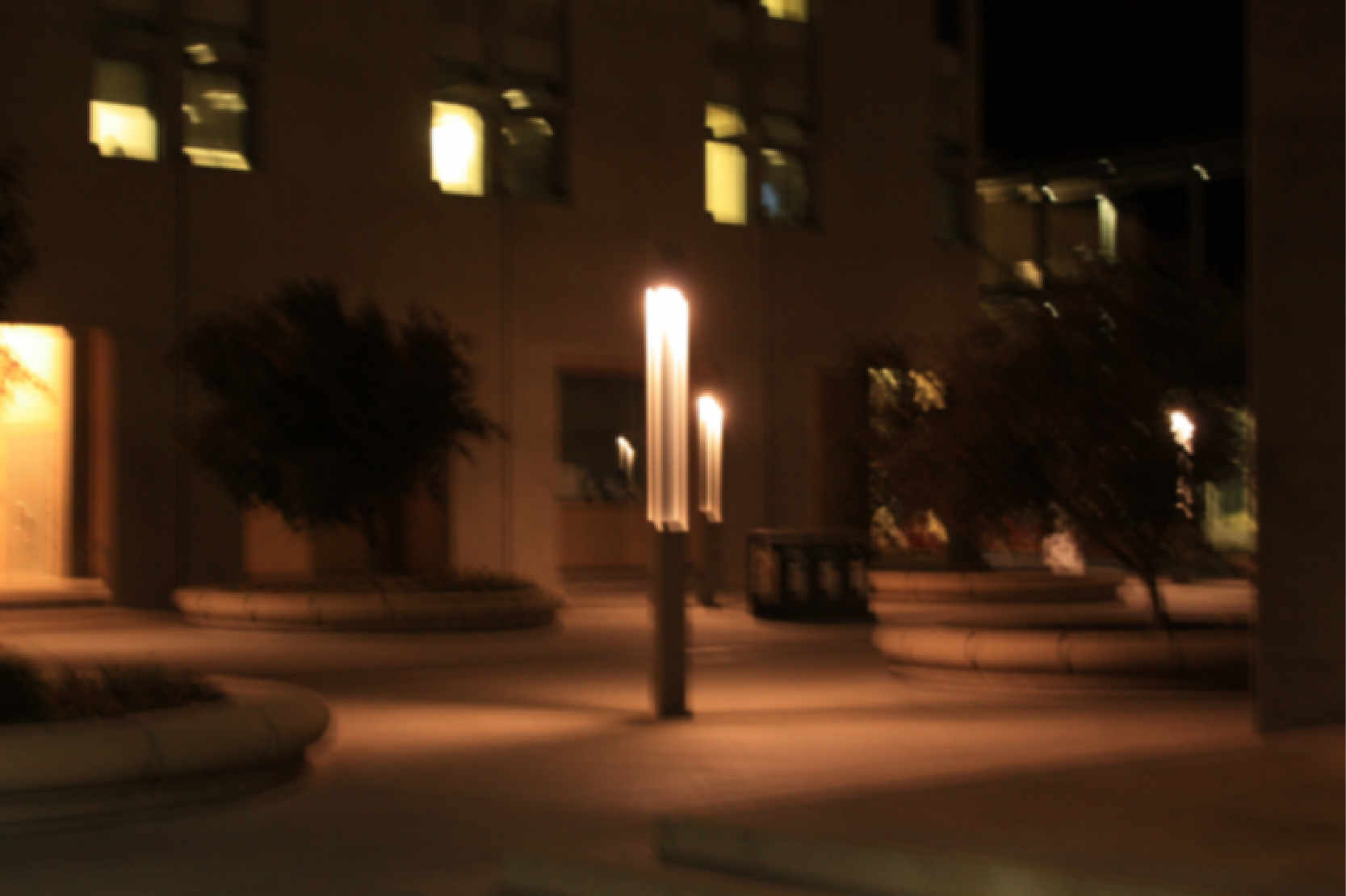} &  \includegraphics[width=0.49\linewidth]{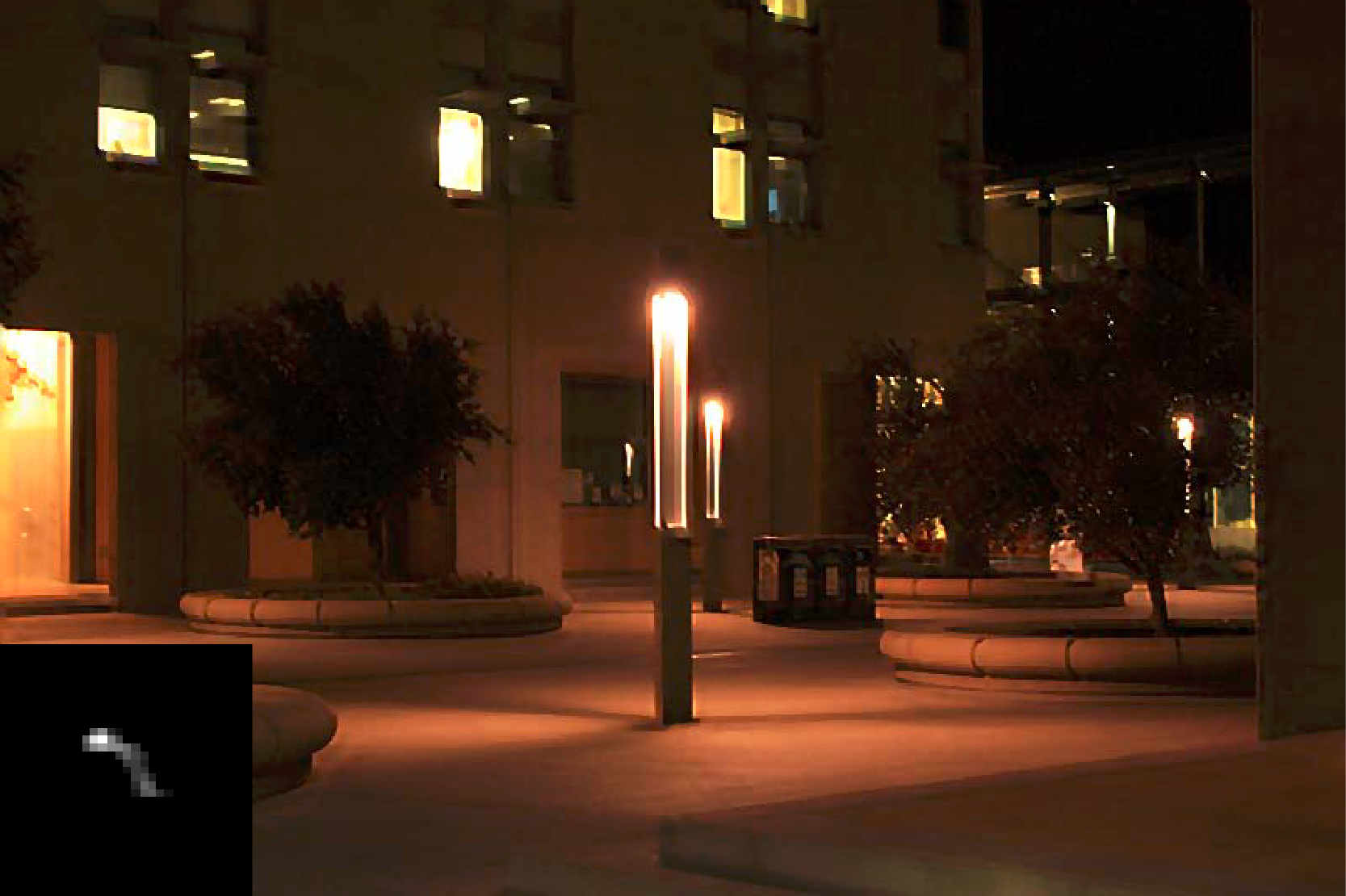}\\
  (a) Blurred &  (b) Hu~\etal~\cite{hu2014deblurring}\\
  \includegraphics[width=0.49\linewidth]{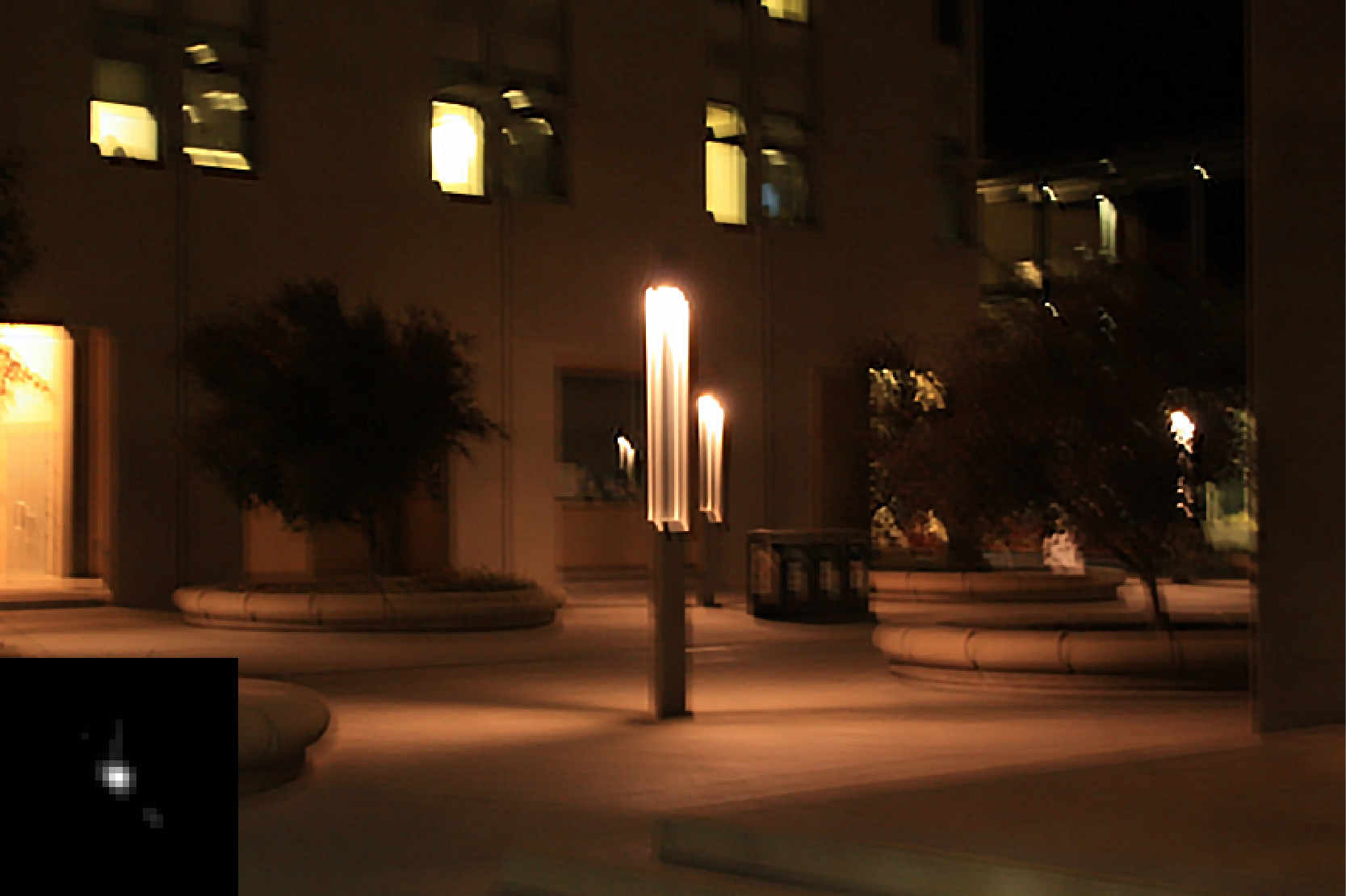} &  \includegraphics[width=0.49\linewidth]{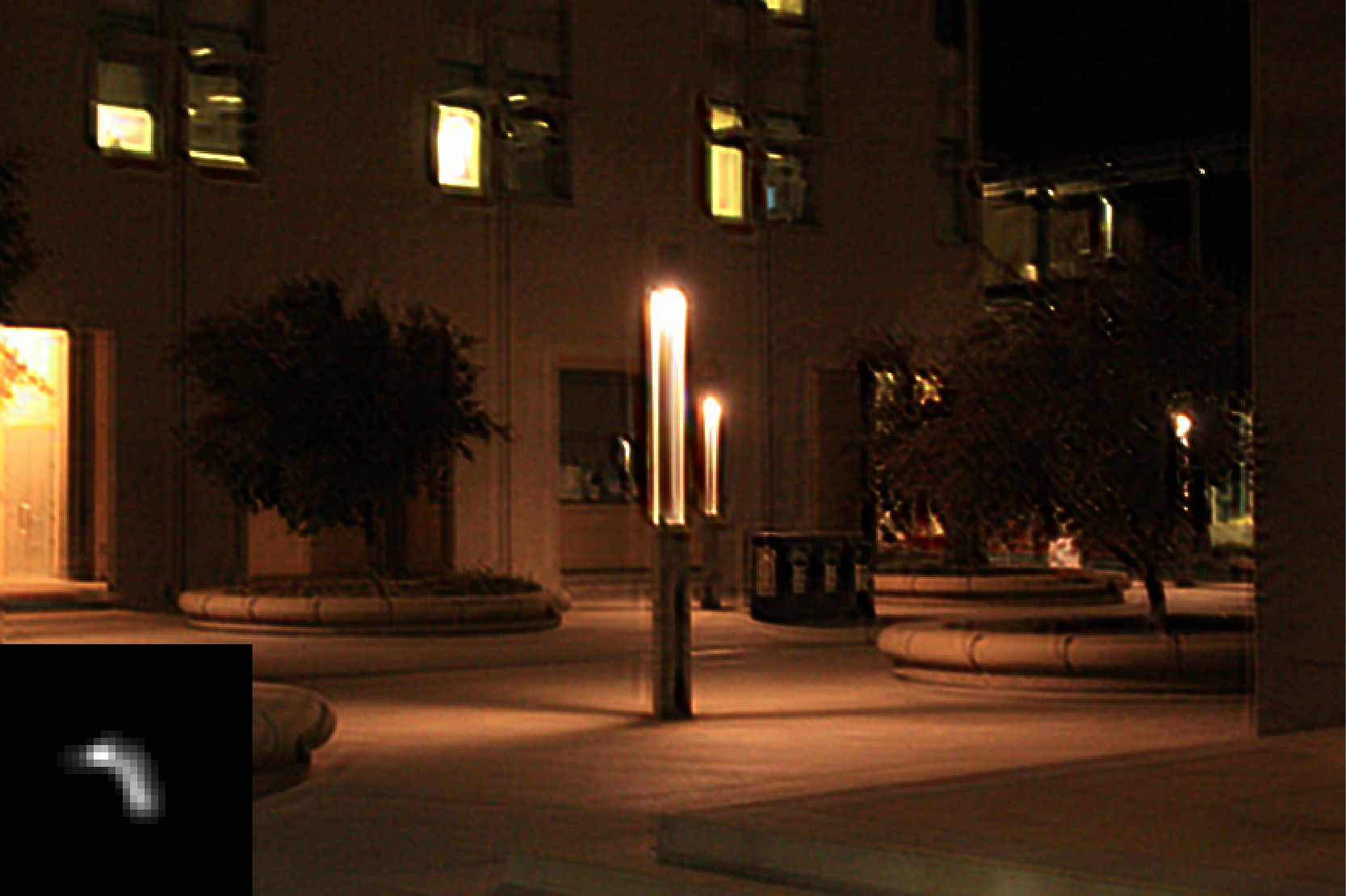} \\
  (c) Xu~\etal~\cite{xu2013unnatural} &  (d) Ours\\
  \end{tabular}
\end{center}
\vspace{-2mm}
\caption{Deblurred results on a low-illumination image. Our method yields comparable results to Hu~\etal~\cite{hu2014deblurring}, which is specially designed for deblurring low-illumination images.}
\label{fig:low}
\end{figure}

Figure~\ref{fig:low} shows an example of the low-illumination image from the dataset of Hu~\etal~\cite{hu2014deblurring}.
Due to the influence of large saturated regions, the natural image deblurring methods fail to generate clear images.
In contrast, our method generates a comparable result with Hu~\etal~\cite{hu2014deblurring}, which is specially designed for the low-illumination images.

\begin{figure}
\footnotesize
\centering
\renewcommand{\tabcolsep}{1pt} 
\renewcommand{\arraystretch}{1} 
\begin{center}
\begin{tabular}{cccc}
  \includegraphics[width=0.24\linewidth]{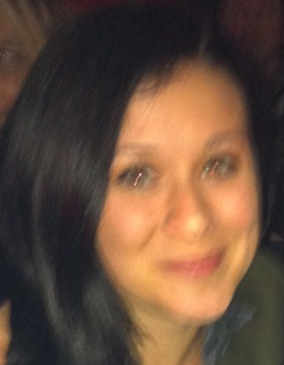} &  \includegraphics[width=0.24\linewidth]{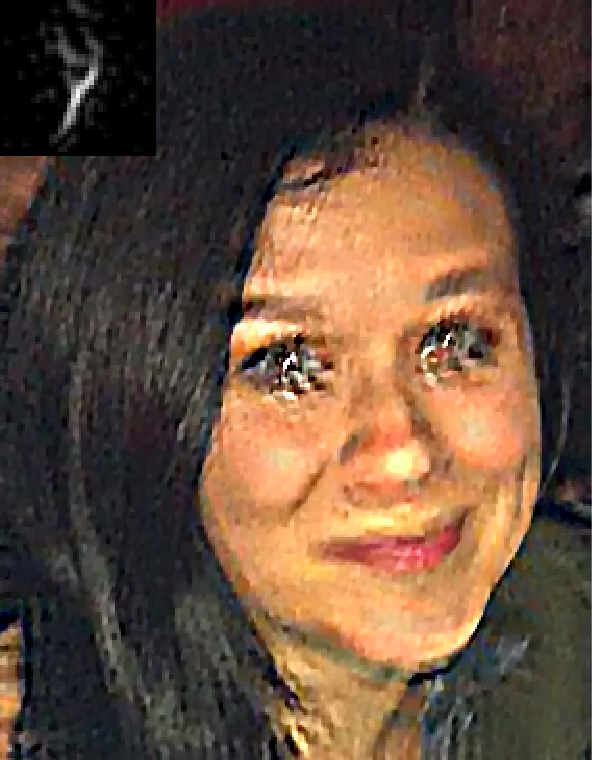} &  \includegraphics[width=0.24\linewidth]{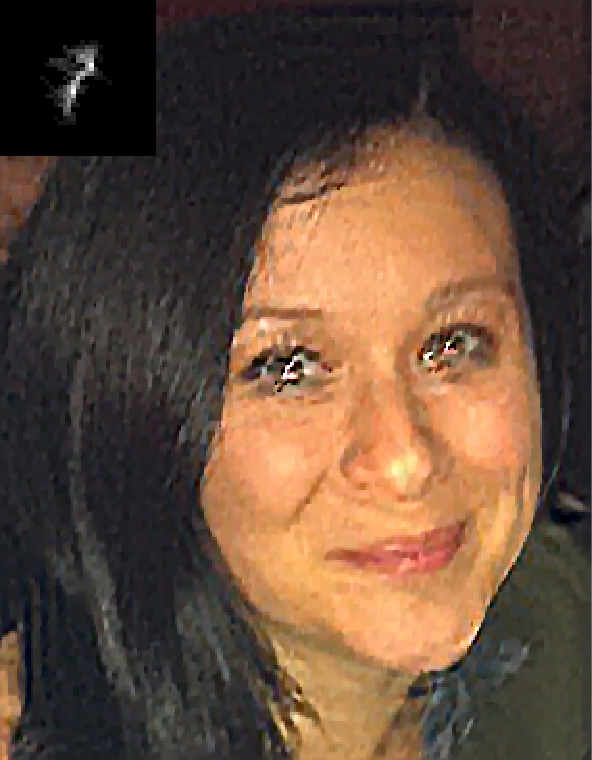} &  \includegraphics[width=0.24\linewidth]{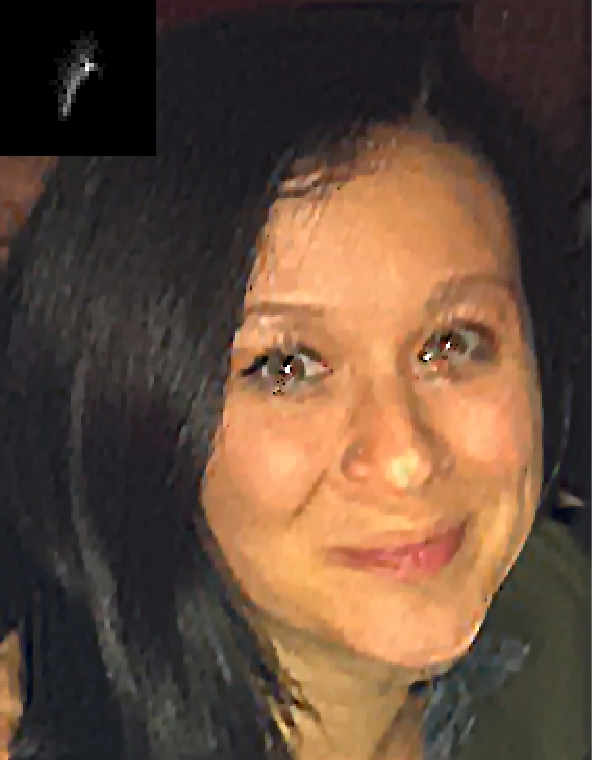} \\
  (a) Blurred image &  (b) Xu~\etal~\cite{xu2013unnatural} &  (c) Yan~\etal~\cite{yan2017image} &  (d) Ours\\
\end{tabular}
\end{center}
\vspace{-2mm}
\caption{Deblurred results on a face image. Our method produces more visually pleasing results.}
\label{fig:face}
\end{figure}

Figure~\ref{fig:face} shows the deblurred results on a face image.
Our result has less ringing artifacts compared with the state-of-the-art methods~\cite{xu2013unnatural, yan2017image}.
We note that the proposed method learns a \emph{generic} image prior but is effective to deblur domain-specific blurred images.

\subsection{Non-uniform deblurring}
\begin{figure}
\footnotesize
\centering
\renewcommand{\tabcolsep}{1pt} 
\renewcommand{\arraystretch}{1} 
\begin{center}
\begin{tabular}{ccc}
  \includegraphics[width=0.32\linewidth]{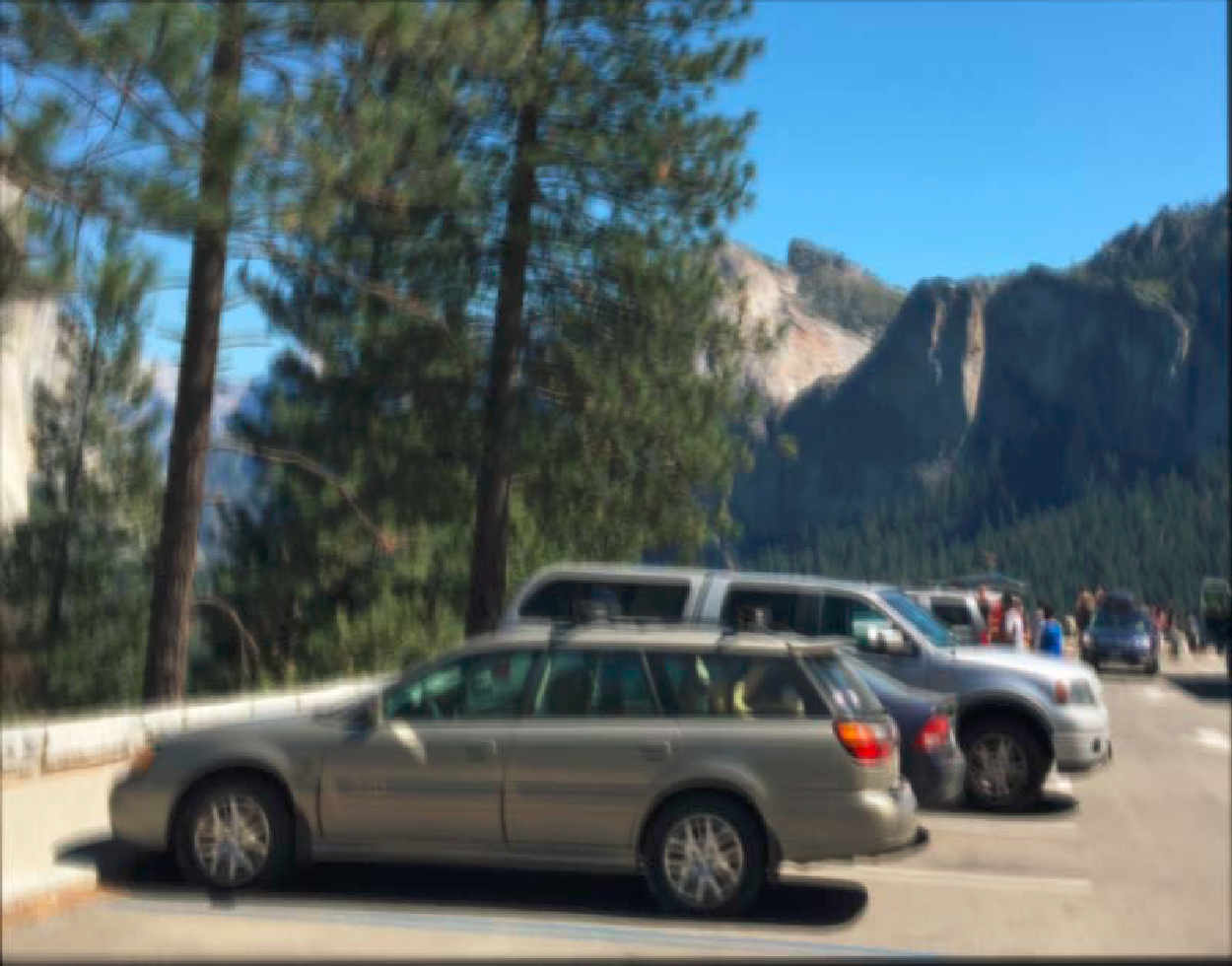} &  \includegraphics[width=0.32\linewidth]{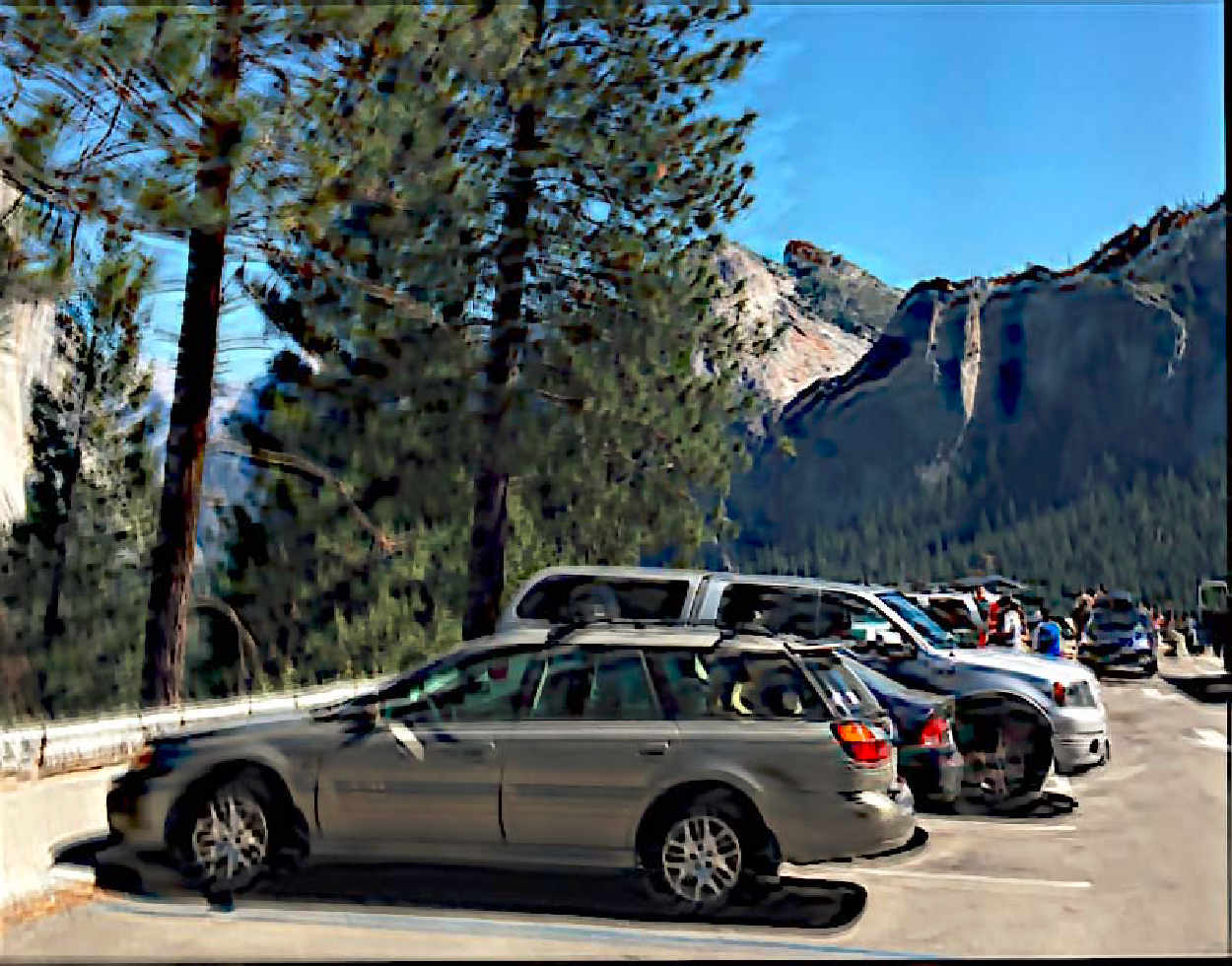} &  \includegraphics[width=0.32\linewidth]{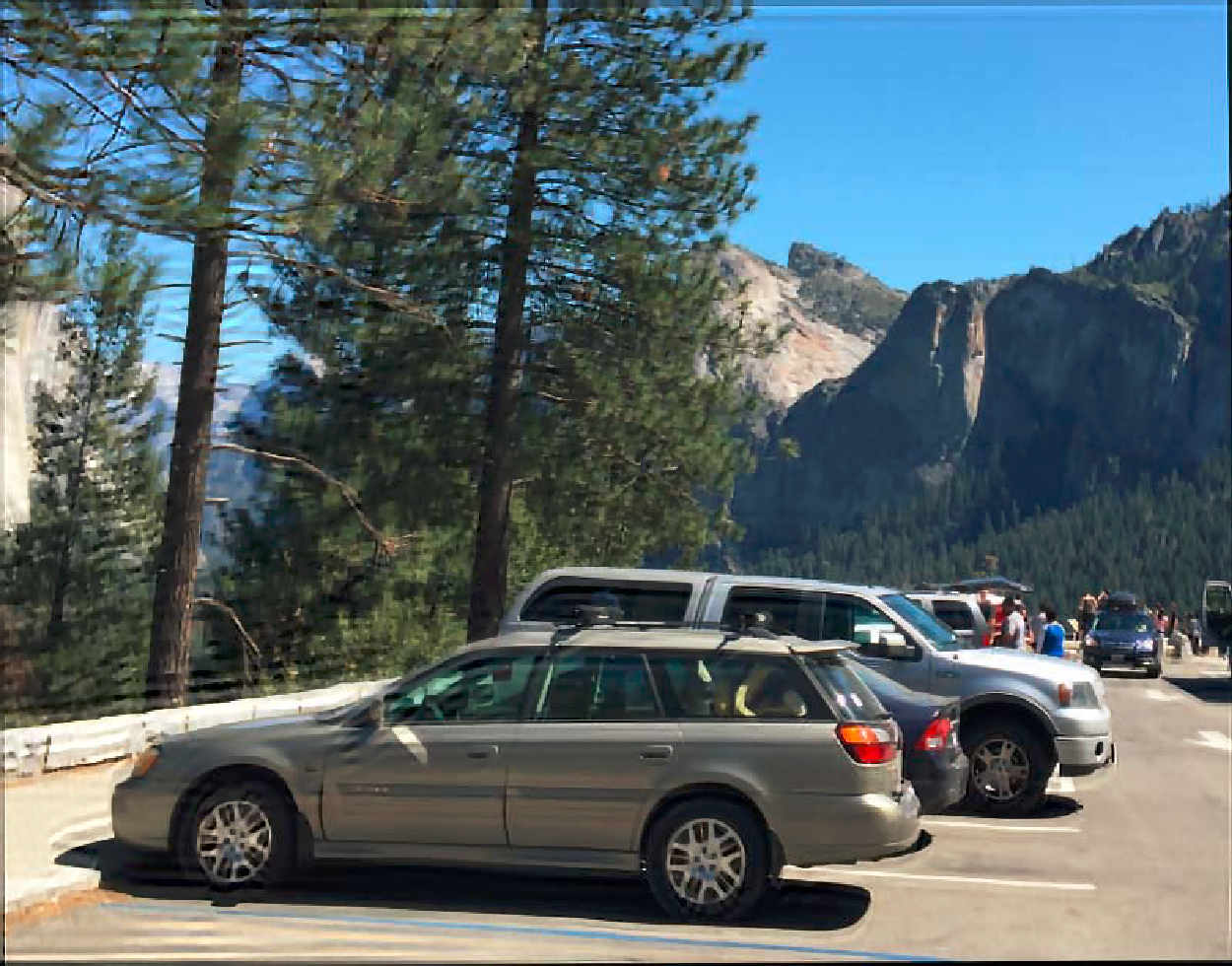}\\
  (a) Blurred image &  (b) Whyte~\etal~\cite{whyte2012non} & (c) Xu~\etal~\cite{xu2013unnatural}\\
  \includegraphics[width=0.32\linewidth]{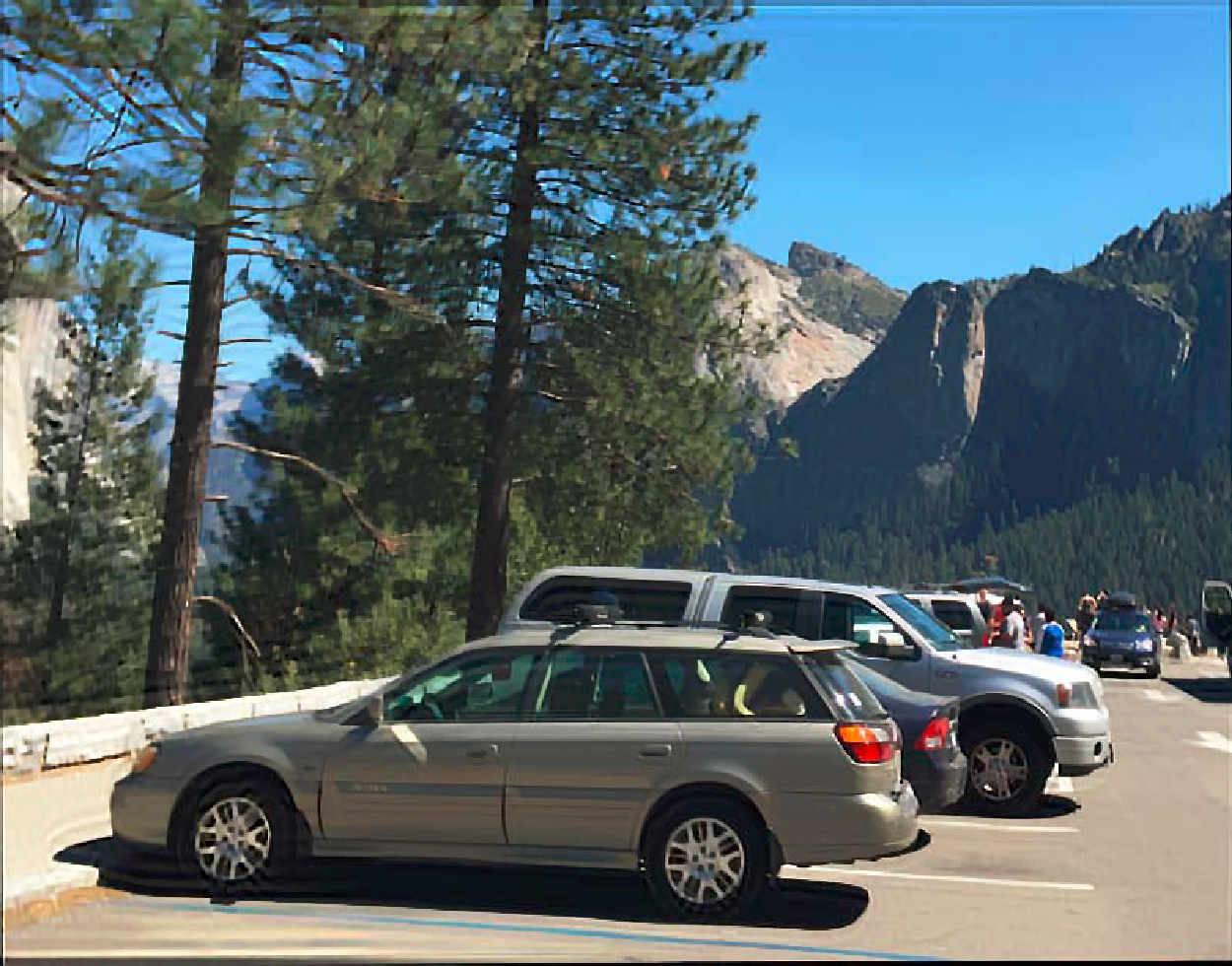} & \includegraphics[width=0.32\linewidth]{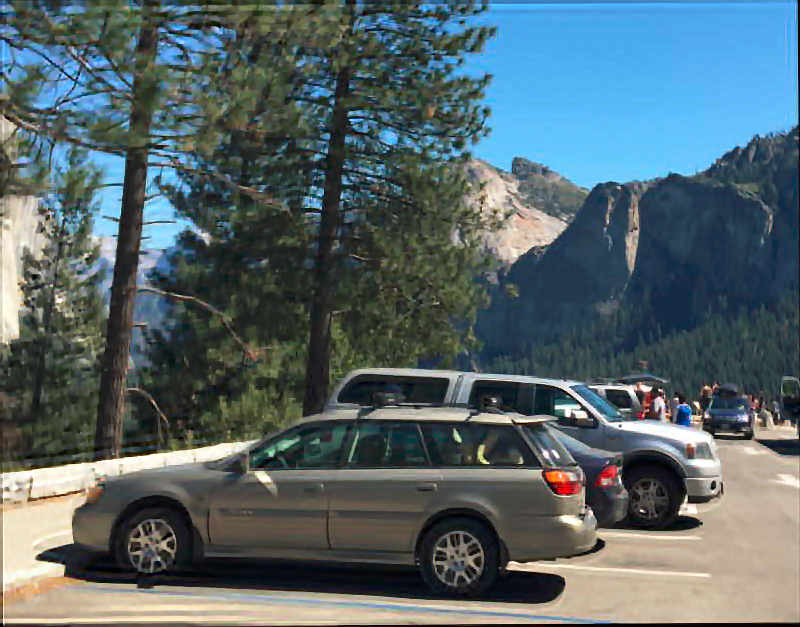} &  \includegraphics[width=0.32\linewidth]{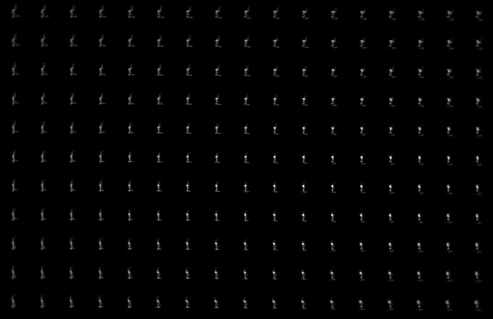}\\
  (d) Pan~\etal~\cite{Pan_2016_CVPR} & (e) Ours &  (d) Our kernels\\
  \end{tabular}
\end{center}
\vspace{-2mm}
\caption{Deblurred results on a real non-uniform blurred image. We extend the proposed method for non-uniform deblurring and provide comparable results with state-of-the-art methods.}
\label{fig:non}
\vspace{-4mm}
\end{figure}

We demonstrate the capability of the proposed method on non-uniform deblurring in Figure~\ref{fig:non}.
Compared with state-of-the-art non-uniform deblurring algorithms~\cite{whyte2012non, xu2013unnatural, Pan_2016_CVPR}, our method produces comparable results with sharp edges and clear textures.

\section{Analysis and Discussion}
\label{sec:AnD}
%
In this section, we analyze the effectiveness of the proposed image prior on distinguishing clear and blurred images, discuss the relation with $L_0$-regularized priors, and analyze the speed, convergence and the limitations of the proposed method.

\subsection{Effectiveness of the proposed image prior}
\begin{figure}[t]
\footnotesize
\centering
\renewcommand{\tabcolsep}{1pt} 
\renewcommand{\arraystretch}{1} 
\begin{center}
\begin{tabular}{cc}
  \includegraphics[width=0.5\linewidth]{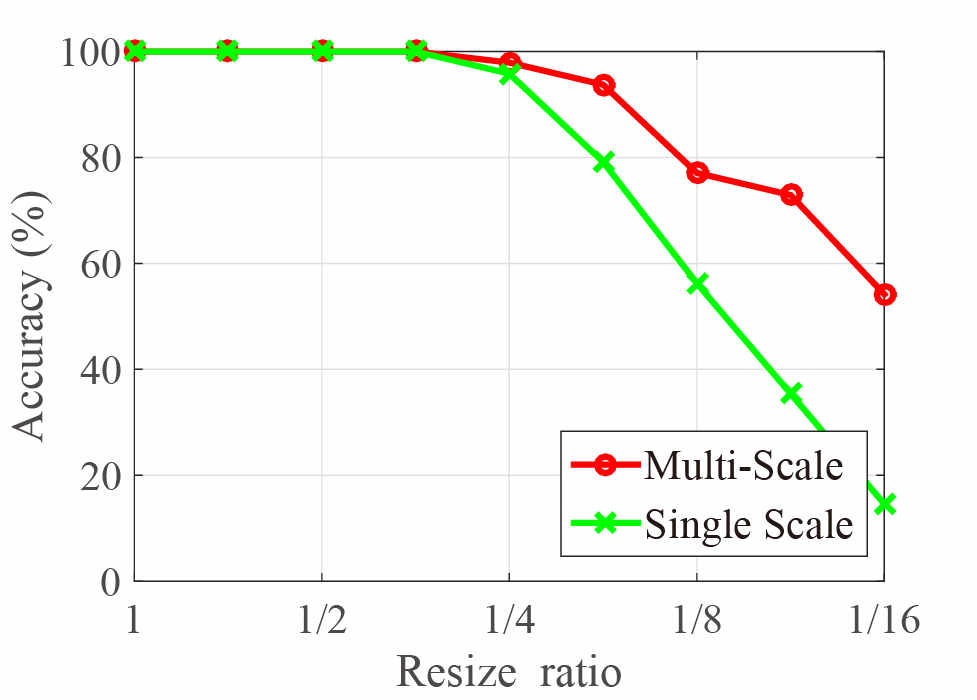} &  \includegraphics[width=0.48\linewidth]{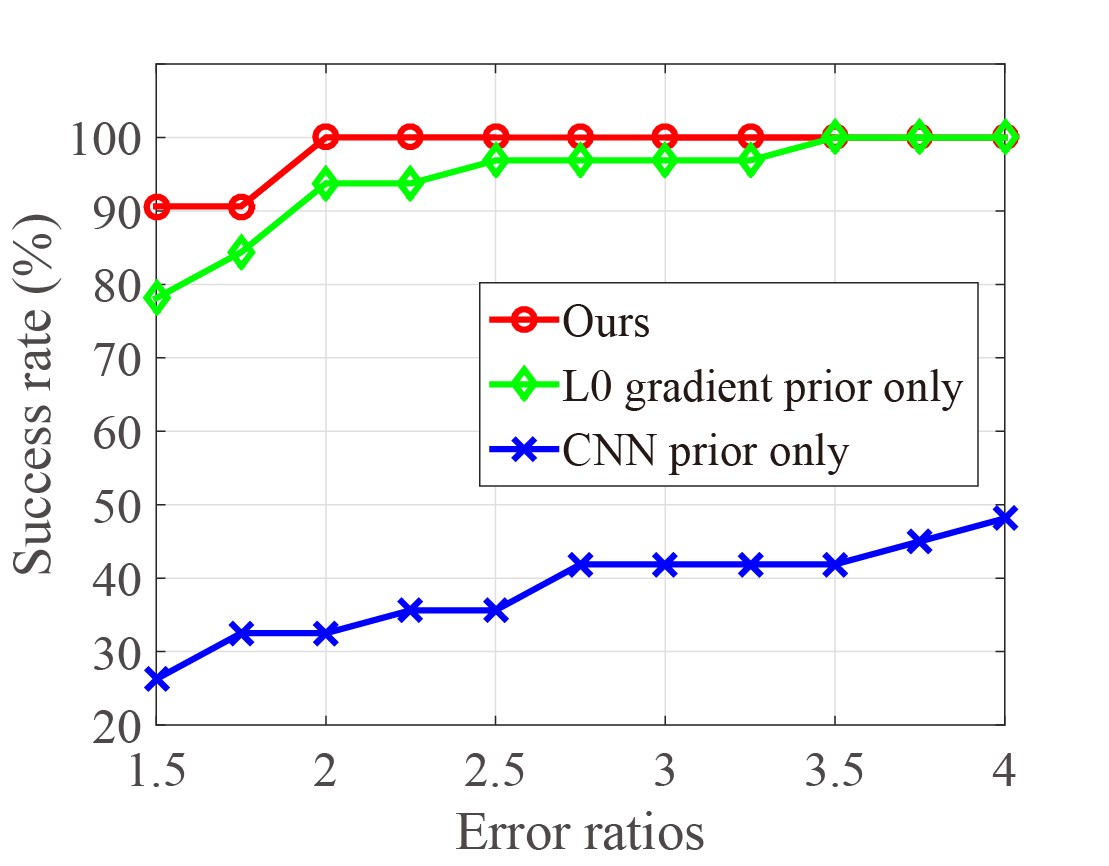}\\
  (a) &  (b) \\
  \end{tabular}
\end{center}
\vspace{-2mm}
\caption{Effectiveness of the proposed CNN prior.
        (a) Classification accuracy on dataset~\cite{Kohler}
        (b) Ablation studies on dataset~\cite{levin2009understanding}
        }
\label{fig:acu}
\end{figure}

We train the binary classification network to predict blurred images as 1 and clear images as 0.
We first use the image size of $200 \times 200$ for training and evaluate the classification accuracy using the images from the dataset of K{\"o}hler~\etal~\cite{Kohler}, where the size of images is $800 \times 800$.
To test the performance of the classifier on different sizes of images, we downsample each image by a ratio between $[1, 1/16]$ and plot the classification accuracy in Figure~\ref{fig:acu} (green curve).
When the size of test images is larger or close to the training image size, the accuracy is near $100\%$.
However, the accuracy drops significantly when images are downscaled by more than $4\times$.
As the downsampling reduces the blur effect, it becomes difficult for the classifier to distinguish blurred and clear images.

To overcome this issue, we adopt a multi-scale training strategy by randomly downsampling each batch of images between $1\times$ and $4\times$.
As shown in the red curve of Figure~\ref{fig:acu}(a), the performance of the classifier becomes more robust to different sizes of input images.
The binary classifier with our multi-scale training strategy is more suitable to be applied in the coarse-to-fine MAP framework.

\begin{figure}
\footnotesize
\centering
\renewcommand{\tabcolsep}{1pt} 
\renewcommand{\arraystretch}{1} 
\begin{center}
\begin{tabular}{ccccc}
  \includegraphics[width=0.23\linewidth]{figure/intro_blur.jpg} &  \includegraphics[width=0.23\linewidth]{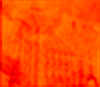}&  \includegraphics[width=0.23\linewidth]{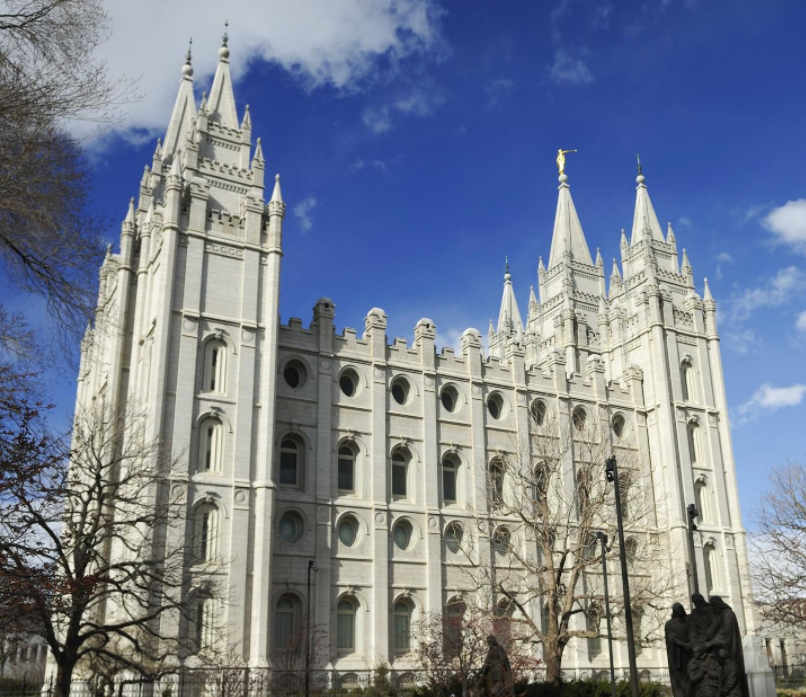}&  \includegraphics[width=0.23\linewidth]{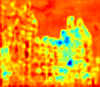} & \includegraphics[height=1.65cm]{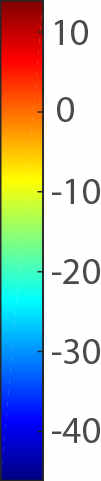}\\
  (a) &  (b) &  (c) & (d) & \\
  \end{tabular}
\end{center}
\vspace{-2mm}
\caption{Activations of a feature map in our binary classification network.
	We show the activation from the C9 layer. The clear image has much lower responses than that of the blurred images.
	(a) Blurred image
	(b) Activation of blurred image
	(c) Clear image
	(d) Activation of clear image}
\label{fig:cnn} 
\vspace{-3mm}
\end{figure}

Figure~\ref{fig:cnn} shows the activation of one feature map from the C9 layer (i.e., the last convolutional layer before the global average pooling) in our classification network.
While the blurred image has a high response on the entire image, the activation of the clear image has a much lower response except for smooth regions, e.g., sky.

\subsection{Relation with $L_0$-regularized priors}
\begin{figure}[t]\footnotesize
\footnotesize
\centering
\renewcommand{\tabcolsep}{1pt} 
\renewcommand{\arraystretch}{1} 
\begin{center}
\begin{tabular}{cccc}
  \includegraphics[width=0.24\linewidth]{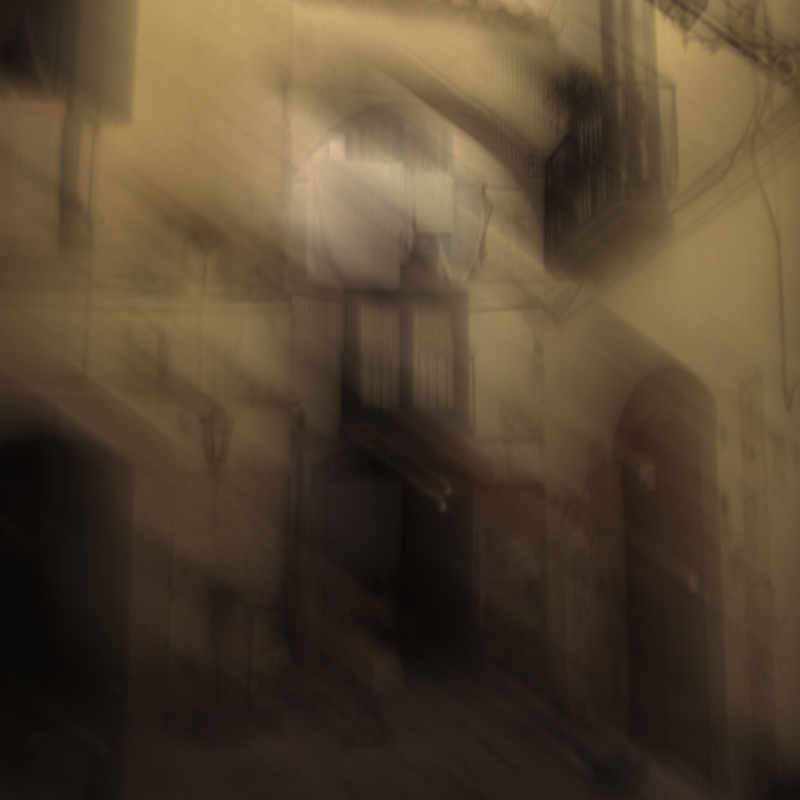} &
  \includegraphics[width=0.24\linewidth]{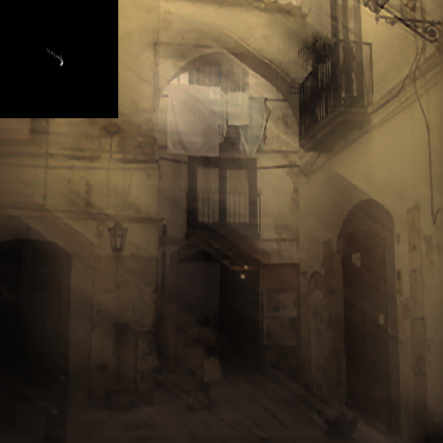} &  \includegraphics[width=0.24\linewidth]{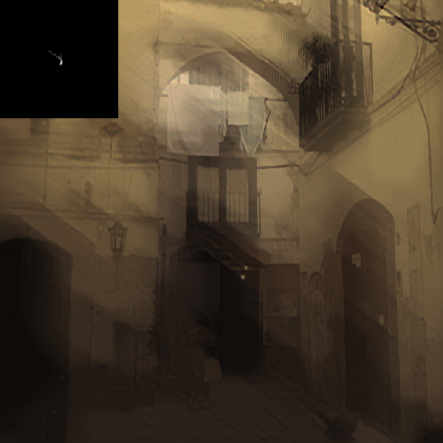} &  \includegraphics[width=0.24\linewidth]{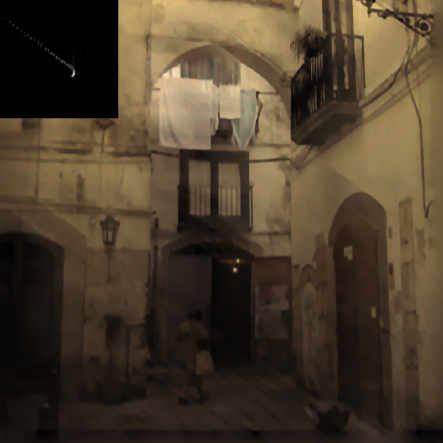} \\
   (a) Input & (b) Pan~\etal~\cite{Pan_2016_CVPR} & (c) Yan~\etal~\cite{yan2017image} & (d) Ours\\
  \includegraphics[width=0.24\linewidth]{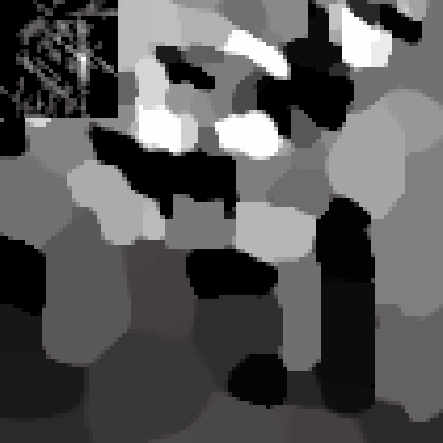} &
  \includegraphics[width=0.24\linewidth]{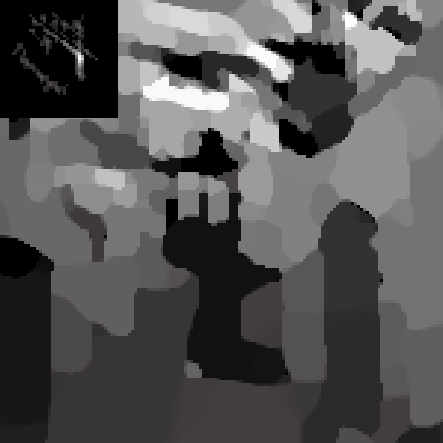} &  \includegraphics[width=0.24\linewidth]{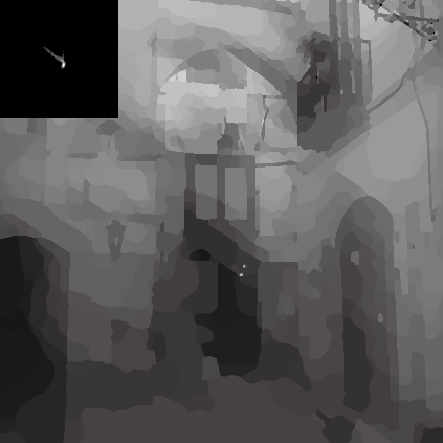} &  \includegraphics[width=0.24\linewidth]{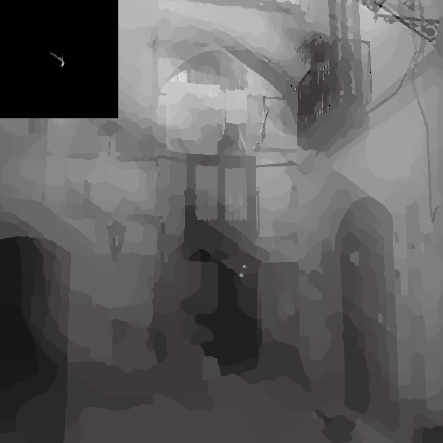} \\
  \multicolumn{4}{c} {(e) Intermediate results of Yan~\etal~\cite{yan2017image}}\\
  \includegraphics[width=0.24\linewidth]{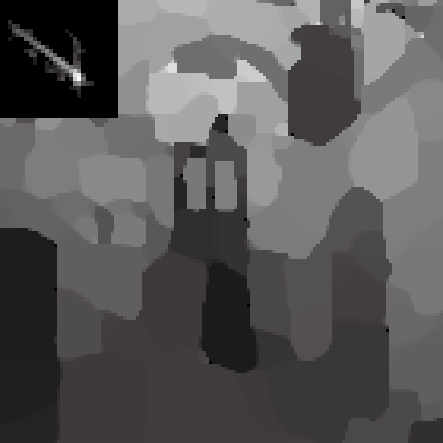} &
  \includegraphics[width=0.24\linewidth]{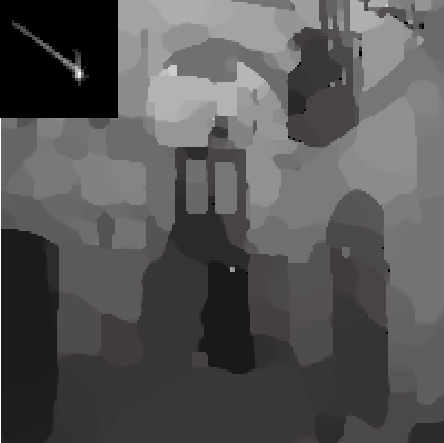} &  \includegraphics[width=0.24\linewidth]{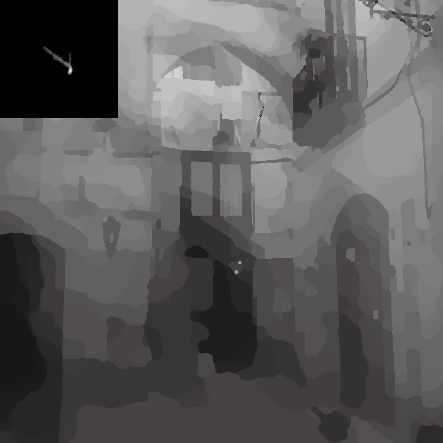} &  \includegraphics[width=0.24\linewidth]{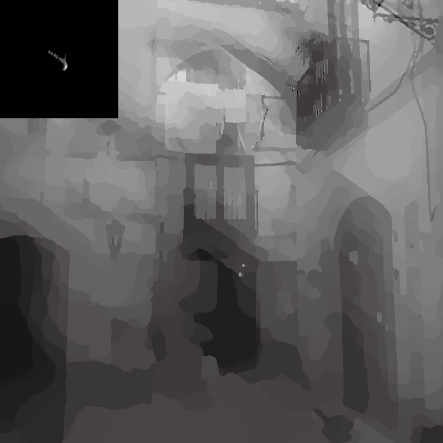} \\
  \multicolumn{4}{c} {(f) Intermediate results of Pan~\etal~\cite{Pan_2016_CVPR}}\\
  \includegraphics[width=0.24\linewidth]{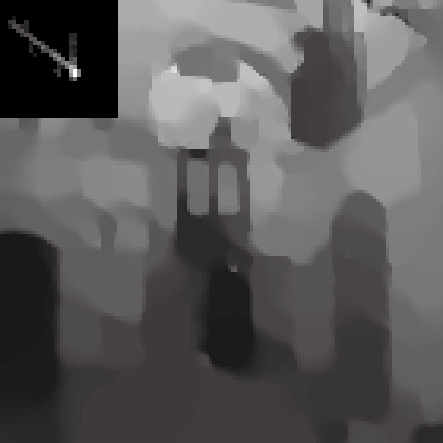} &
  \includegraphics[width=0.24\linewidth]{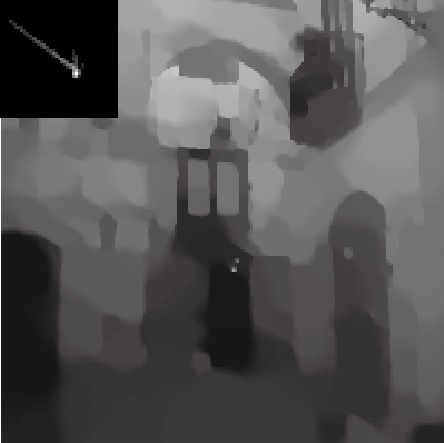} &  \includegraphics[width=0.24\linewidth]{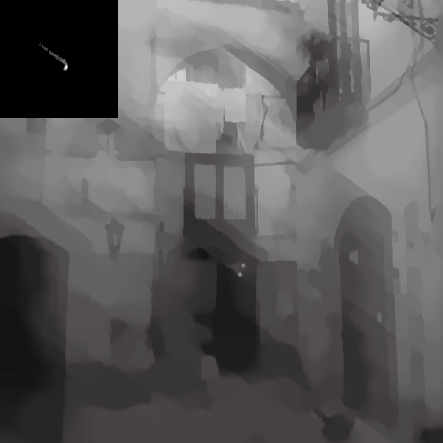} &  \includegraphics[width=0.24\linewidth]{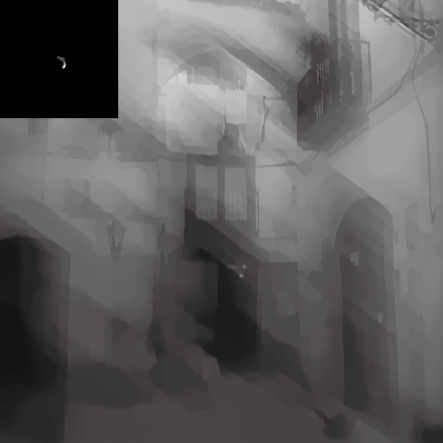} \\
  \multicolumn{4}{c} {(g) Intermediate results of our method without using discriminative prior}\\
  \includegraphics[width=0.24\linewidth]{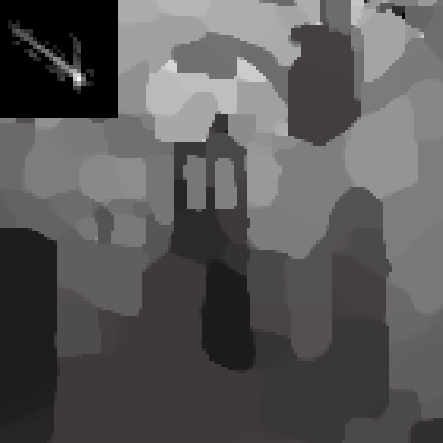} &
  \includegraphics[width=0.24\linewidth]{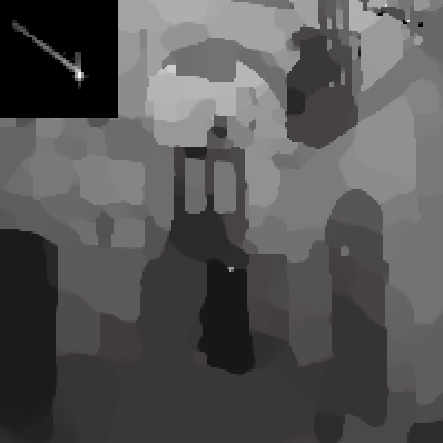} &  \includegraphics[width=0.24\linewidth]{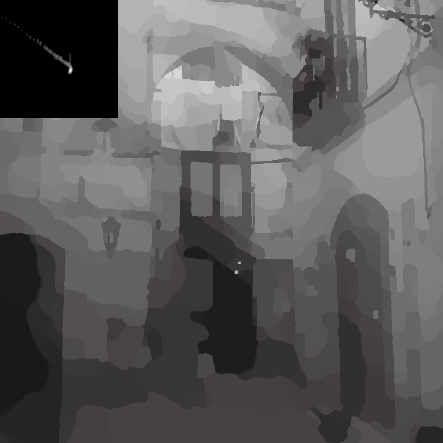} &  \includegraphics[width=0.24\linewidth]{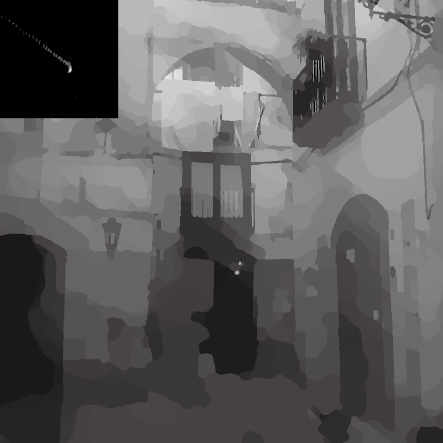} \\
  \multicolumn{4}{c} {(h) Intermediate results of our method using discriminative prior}\\
\end{tabular}
\end{center}
\vspace{-2mm}
\caption{Deblurred and intermediate results.
	We compare the deblurred results with state-of-the-art methods~\cite{yan2017image,Pan_2016_CVPR} in (a)-(d) and illustrate the intermediate latent images over iterations (from left to right) in (e)-(h).
	Our discriminative prior recovers intermediate results with more strong edges for kernel estimation. }
\label{fig:interm} 
\end{figure}

Several methods~\cite{Pan2014Deblurring,xu2013unnatural} adopt the $L_0$-regularized priors in blind image deblurring due to the strong sparsity of the $L_0$ norm.
State-of-the-art approaches~\cite{Pan_2016_CVPR, yan2017image} enforce the $L_0$ sparsity on the extreme channels (i.e., dark and bright channels) as the blur process affects the distribution of the extreme channels.
The proposed approach also includes the $L_0$ gradient prior for regularization.
The intermediate results in Figure~\ref{fig:interm} show that the methods based on $L_0$-regularized prior on extreme channels~\cite{Pan_2016_CVPR, yan2017image} fail to recover strong edges when there are not enough dark or bright pixels.
Figure~\ref{fig:interm}(g) shows that the proposed method without the learned discriminative image prior (i.e., use $L_0$ gradient prior only) cannot well reconstruct strong edges for estimating the blur kernel.
In contrast, our discriminative image prior restores more sharp edges in the early stage of the optimization and improve the blur kernel estimation.

To better understand the effectiveness of each term in (\ref{4.1}), we conduct an ablation study on the dataset of Levin~\etal~\cite{levin2009understanding}.
As shown in Figure~\ref{fig:acu}(b), while the $L_0$ gradient prior helps to preserve more image structures, the integration with the proposed CNN prior leads to state-of-the-art performance.

\subsection{Runtime and convergence property}
\begin{figure}
\footnotesize
\centering
\begin{center}
\begin{tabular}{cc}
  \includegraphics[width=0.45\linewidth]{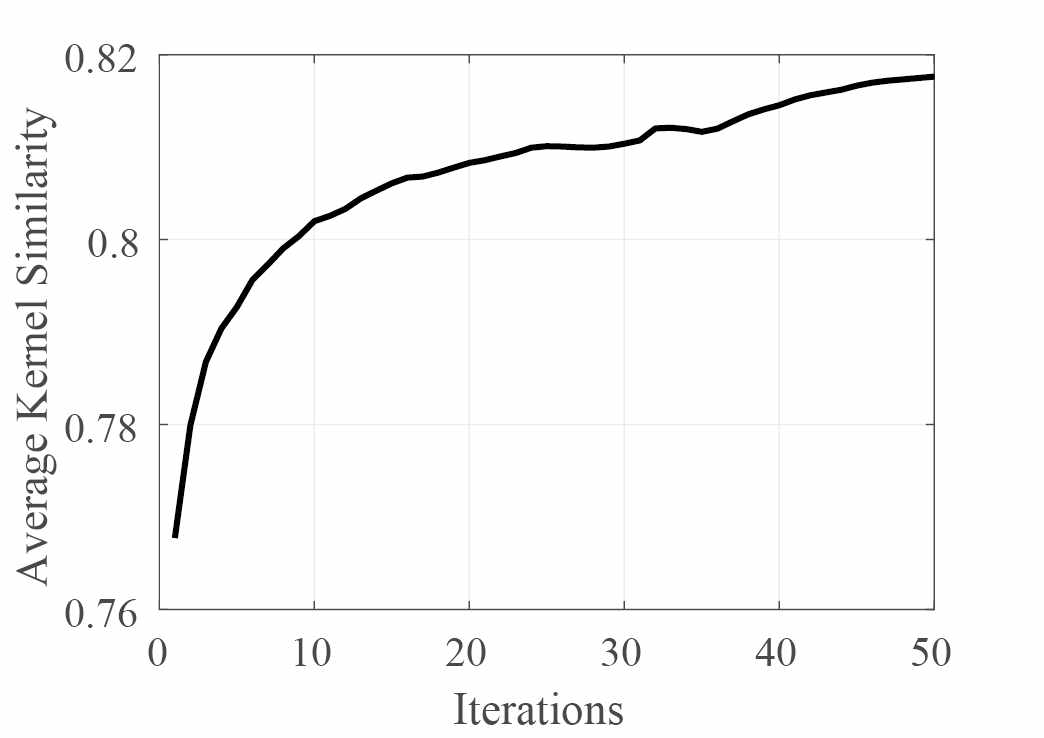} & \hspace{-0.45cm} \includegraphics[width=0.45\linewidth]{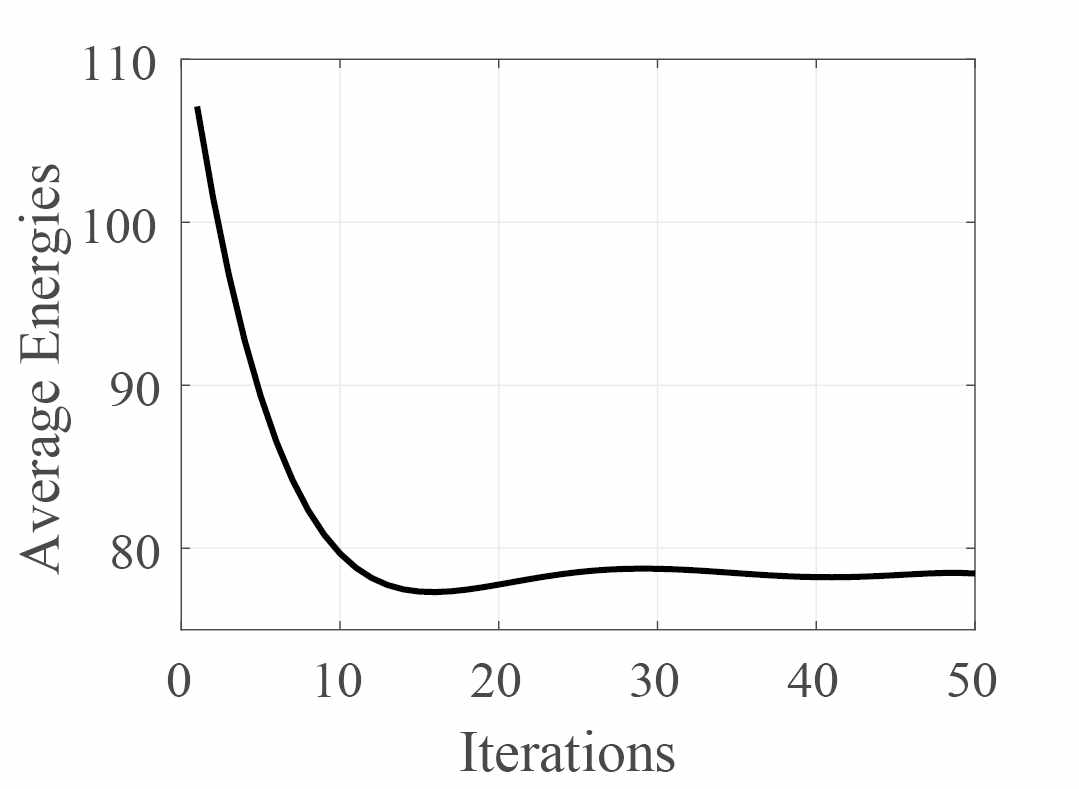}\\
  (a) Kernel similarity & \hspace{-0.45cm} (b) Energy function\\
  \end{tabular}
\end{center}
\vspace{-2mm}
\caption{
	Convergence analysis of the proposed optimization method.
	We analyze the kernel similarity~\cite{hu2012good} and the objective function~\eqref{4.1} at the finest image scale.
	Our method converges well within 50 iterations.}
\label{fig:conver}
\vspace{-4mm}
\end{figure}

Our algorithm is based on the efficient half-quadratic splitting and gradient decent methods.
We test the state-of-the-art methods on different sizes of images and report the average runtime in Table~\ref{tab:runtime}.
The proposed method runs competitively with state-of-the-art approaches~\cite{Pan_2016_CVPR, yan2017image}.
In addition, we quantitatively evaluate convergence of the proposed optimization method using images from the dataset of Levin~\etal~\cite{levin2009understanding}.
We compute the average kernel similarity~\cite{hu2012good} and the values of the objective function~\eqref{4.1} at the finest image scale.
Figure~\ref{fig:conver} shows that our algorithm converges well whithin 50 iterations.

\begin{table}[!t]
  \centering
  \scriptsize
  \caption{Runtime comparisons. We report the average runtime (seconds) on three different sizes of images.}
  \label{tab:runtime}
  \begin{tabular}{c|ccc}
    \toprule
    Method & $255\times255$ & $600\times600$ & $800\times800$ \\
    \midrule
    Xu~\etal~\cite{xu2013unnatural} (C++) & 1.11 & 3.56 & 4.31 \\
    Krishnan~\etal~\cite{krishnan2011blind} (MATLAB) & 24.23 & 111.09 & 226.58 \\
    Levin~\etal~\cite{levin2011efficient} (MATLAB) & 117.06 & 481.48 & 917.84 \\
    Pan~\etal~\cite{Pan_2016_CVPR} (MATLAB) & 134.31 & 691.71 & 964.90 \\
    Yan~\etal~\cite{yan2017image} (MATLAB) & 264.78 & 996.03 & 1150.48 \\
    Ours (MATLAB) & 109.27 & 379.52 & 654.65 \\
    \bottomrule
  \end{tabular}
  \vspace{-3mm}
\end{table}

\subsection{Limitations}
\begin{figure}
\footnotesize
\centering
\renewcommand{\tabcolsep}{1pt} 
\renewcommand{\arraystretch}{1} 
\begin{center}
\begin{tabular}{ccc}
  \includegraphics[width=0.32\linewidth]{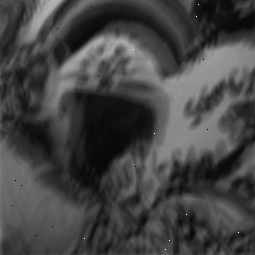} &  \includegraphics[width=0.32\linewidth]{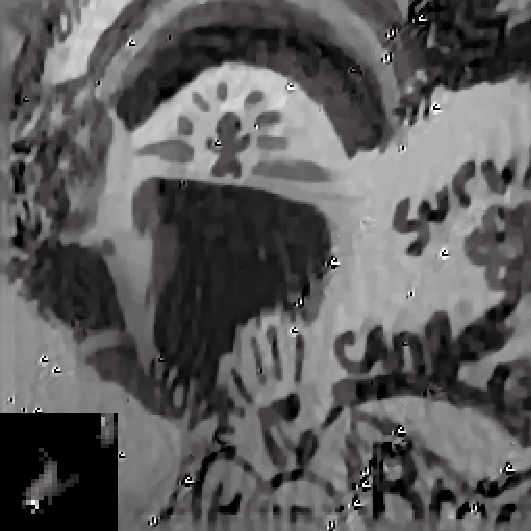} &
  \includegraphics[width=0.32\linewidth]{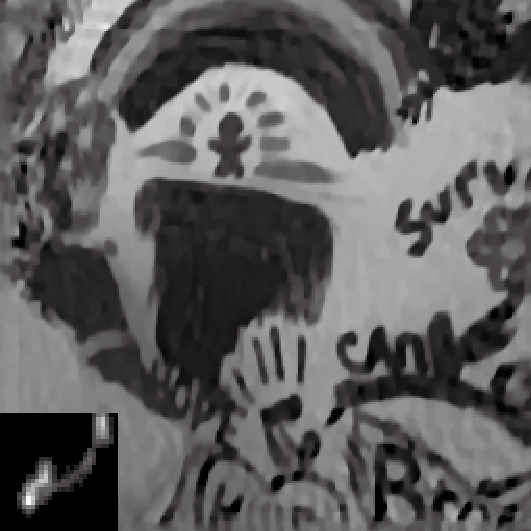}\\
  (a) &  (b) &  (c)\\
  \end{tabular}
\end{center}
\vspace{-2mm}
\caption{Limitations of the proposed method.
	Our learned image prior is not effective on handling images with salt and pepper noise.
	(a) Blurred image
	(b) Our deblurred results
	(c) Our deblurred results by first applying a median filter on blurred image.}
\label{fig:noise}
\vspace{-4mm}
\end{figure}

As our classification network is trained on image intensity, the learned image prior might be less effective when input images contain significant noise and outliers.
Figure~\ref{fig:noise} shows an example with salt and pepper noise in the input blurred image.
In this case, our classification network cannot differentiate the blurred image ($f(B) \simeq 0$) due to the influence of salt and pepper noise.
Therefore, the proposed prior cannot restore the image well as shown in Figure~\ref{fig:noise}(b).
A simple solution is to first apply a median filter on the input image before adopting our approach for deblurring.
As shown in Figure~\ref{fig:noise}(c), although we can reconstruct a better deblurred result, the details of the recovered images are not preserved well.
Future work will consider joint deblurring and denoising in a principal way.

\section{Conclusions}
\label{sec:conclusions}
In this paper, we propose a data-driven discriminative prior for blind image deblurring.
We learn the image prior via a binary classification network based on a simple criterion: the prior should favor clear images over blurred images on various of scenarios.
We adopt a global average pooling layer and a multi-scale training strategy to make the network more robust to different sizes of images.
We then embed the learned image prior into a coarse-to-fine MAP framework and develop an efficient half-quadratic splitting algorithm for blur kernel estimation.
Our prior is effective on several types of images, including natural, text, face and low-illumination images, and can be easily extended to handle non-uniform deblurring.
Extensive quantitative and qualitative comparisons demonstrate that the proposed method performs favorably against state-of-the-art generic and domain-specific blind deblurring algorithms.

\section*{Acknowledgements}
This work is partially supported by NSFC (No. 61433007, 61401170, and 61571207), NSF CARRER (No. 1149783), and gifts from Adobe and Nvidia.
Li L. is supported by a scholarship from China Scholarship Council.

\newpage
{\small
\bibliographystyle{ieee}
\bibliography{reference}
}
\end{document}